\documentclass[english]{article}
\usepackage[OT1]{fontenc}
\usepackage[latin9]{inputenc}
\usepackage{geometry}
\usepackage{color}
\usepackage{babel}
\usepackage{float}
\usepackage{mathtools}
\usepackage{bm}
\usepackage{dsfont}
\usepackage{amsmath}
\usepackage{amssymb}

\usepackage[unicode=true,pdfusetitle,
 bookmarks=true,bookmarksnumbered=false,bookmarksopen=false,
 breaklinks=false,pdfborder={0 0 1},backref=false,colorlinks=true]
 {hyperref}

\usepackage{authblk}
\usepackage{nicefrac}
\usepackage{subcaption}
\makeatletter

\floatstyle{ruled}
\newfloat{algorithm}{tbp}{loa}
\providecommand{\algorithmname}{Algorithm}
\floatname{algorithm}{\protect\algorithmname}

\usepackage{babel}

\usepackage{algorithmic}

\usepackage{amsthm}

\theoremstyle{plain}

\newtheorem{lemma}{\textbf{Lemma}}
\newtheorem{theorem}{\textbf{Theorem}}
\newtheorem{corollary}{\textbf{Corollary}}

\newtheorem{definition}{\textbf{Definition}}
\newtheorem{fact}{\textbf{Fact}}
\newtheorem{proposition}{\textbf{Proposition}}

\theoremstyle{definition}

\newtheorem{claim}{\textbf{Claim}}

\usepackage{color}
\definecolor{cm}{RGB}{0,0,200}

\definecolor{yy}{RGB}{0,200,0}

\def\R{\mathbb{R}}

\newcommand{\cE}{\mathcal{E}}
\newcommand{\cF}{\mathcal{F}}
\newcommand{\cG}{\mathcal{G}}
\newcommand{\defeq}{\coloneqq}

\makeatother

\begin{document}
\title{Top-$K$ ranking with a monotone adversary}
\author[1]{Yuepeng Yang}
\author[2]{Antares Chen}
\author[2]{Lorenzo Orecchia}
\author[1]{Cong Ma}
\affil[1]{Department of Statistics, University of Chicago}
\affil[2]{Department of Computer Science, University of Chicago}
\maketitle

\begin{abstract}
In this paper, we address the top-$K$ ranking problem with a monotone adversary.
We consider the scenario where a comparison graph is randomly generated and the adversary is allowed to add arbitrary edges. 
The statistician's goal is then to accurately identify the top-$K$ preferred items based on pairwise comparisons derived from this semi-random comparison graph. 
The main contribution of this paper is 
to develop a weighted maximum likelihood estimator (MLE) that achieves near-optimal sample complexity, up to a $\log^2(n)$ factor, where $n$ denotes the number of items under comparison.
This is made possible through a combination of analytical and algorithmic innovations. 
On the analytical front, we provide a refined~$\ell_\infty$ error analysis of the weighted MLE that is more explicit and tighter than existing analyses. It relates the~$\ell_\infty$ error with the spectral properties of the weighted comparison graph. Motivated by this, our algorithmic innovation involves the development of an SDP-based approach to reweight the semi-random graph and meet specified spectral properties. 
Additionally, we propose a first-order method based on the Matrix Multiplicative Weight Update (MMWU) framework. This method efficiently solves the resulting SDP in nearly-linear time relative to the size of the semi-random comparison graph.
\end{abstract}

\section{Introduction}

In this paper, we consider the problem of ranking $n$ items given pairwise comparisons among them.
This problem finds numerous applications in recommendation system \cite{wang2018streaming},
rating players \cite{elo1967proposed}, web search~\cite{dwork2001rank}, etc. One widely adopted model
for pairwise comparison data is the Bradley-Terry-Luce (BTL) model
\cite{bradley1952rank,luce2005individual}.  In this model, one assumes a latent score
vector $\bm{\theta}^{\star}\in\mathbb{R}^{n}$, and that the Bernoulli
outcome of the comparison between items $i$ and $j$ follows 
\[
\mathbb{P}\left[\text{\text{ item \ensuremath{i} is preferred over item \ensuremath{j} }}\right]=\frac{e^{\theta_{i}^{\star}}}{e^{\theta_{i}^{\star}}+e^{\theta_{j}^{\star}}}.
\]
It is intuitive that under the BTL model, a higher score indicates a higher chance of winning a comparison. 

In practice, the comparisons are often made for a subset of all possible pairs. A popular model
to accommodate this situation is the uniform sampling model~\cite{chen2015spectral,chen2019spectral}, i.e., each pair is compared independently
with probability~$p$. Uniform sampling is quite convenient for theory. An an example, under this sampling mechanism, \cite{chen2019spectral} shows that with high probability, 
the (regularized) maximum likelihood estimator (MLE) \cite{ford1957solution} exactly identifies
the top $K$ preferred items with an optimal sample complexity:
\begin{align}\label{eq:optimal-sample-complexity-intro}
n^2 p\gtrsim\frac{n \log(n)}{\Delta_{K}^{2}},    
\end{align}
where $\Delta_{K}$ measures the latent score difference between the $K$-th
and the $(K+1)$-th preferred items, and $n^2 p$ is the expected number of comparisons. 

Uniform sampling, while convenient for theoretical purposes, is often too ideal to match practice. This motivates a line of work~\cite{shah2016estimation, li2022ell_,chen2023ranking} that goes beyond uniform sampling and focuses on general sampling mechanisms. However, the theoretical guarantees are far from satisfactory. Take the recent work \cite{li2022ell_} on the general
sampling case as an example. The (regularized) MLE requires a sample complexity of $n^2 p\gtrsim\Delta_{K}^{-2} p^{-1} n \log(n)$
when applied to the special case of uniform sampling. 
For a sparse random graph, i.e., when $p \gtrsim \log(n)/n$, this sample complexity could be $n$ times larger than the optimal one~\eqref{eq:optimal-sample-complexity-intro}.

\begin{figure}[t]
\begin{minipage}[t]{.45\linewidth}
    \centering
    \includegraphics[height=10em]{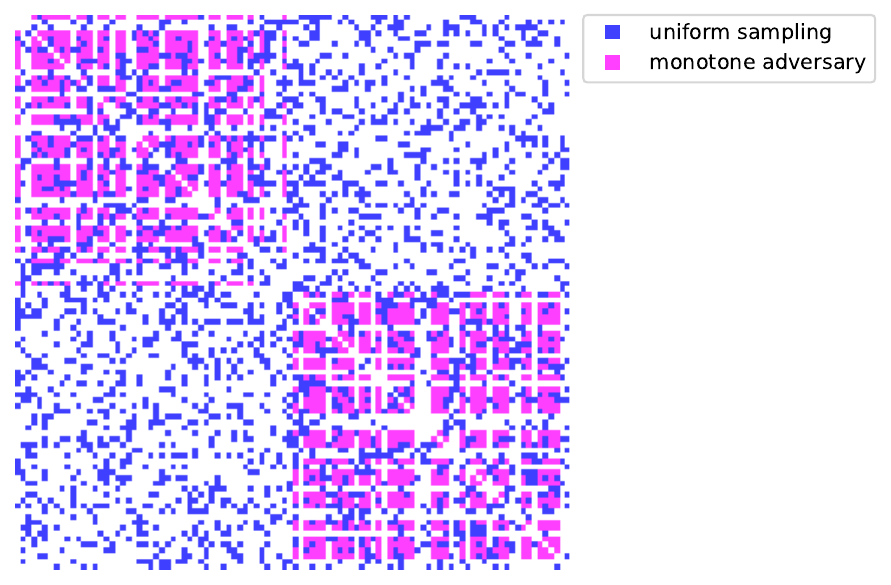}
    \caption{Adjacency matrix of a semi-random comparison graph. Each non-white square corresponds to a non-zero entry in the adjacency matrix.}
    \label{fig:semirandom}
\end{minipage}%
\hspace{0.05\linewidth}
\begin{minipage}[t]{.45\linewidth}
    \centering
    \includegraphics[height=10em]{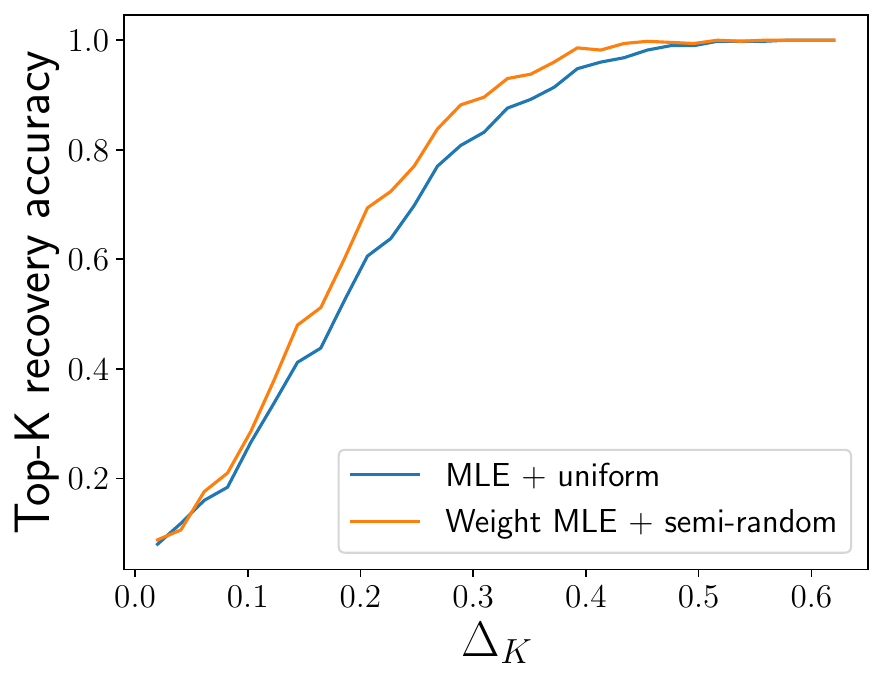}
    \caption{Accuracy of top-$K$ recovery for MLE under uniform sampling and weighted MLE under semi-random sampling. See Section~\ref{app:experiments} for the experiment setup.}
    \label{fig:topK}
\end{minipage}

\label{fig:semirandom_topK}
\end{figure}

In this paper, we aim to find a middle ground between uniform and general
sampling mechanisms. Inspired by a line of work \cite{blum1995coloring,feige2001heuristics,makarychev2012approximation,moitra2016robust, awasthi2018clustering,cheng2018non,kelner2023semi,gao2023robust},
we consider the so-called \emph{semi-random sampling} where some benign
adversary is allowed to make more comparisons in addition to uniform
sampling; see Figure~\ref{fig:semirandom}. Given its monotone nature, this is sometimes also referred to
as a \emph{monotone adversary}.\footnote{In this paper, we use the terms semi-random and monotone interchangeably.}  An important example of semi-random sampling is non-uniform sampling, where each pair $(i, j)$ is sampled with non-uniform probability $p_{ij} \in [p, 1]$. Clearly, this is more flexible and practical than uniform sampling.

Intuitively, the monotone adversary should bring no harm
to the ranking problem as it only reveals more information about
the underlying score vector $\bm{\theta}^{\star}$. However, it is well documented in the literature that the monotone adversary poses serious
algorithmic and analytical challenges for a variety of problems. In community detection, \cite{moitra2016robust} shows that the information-theoretic detection limit could shift given a monotone adversary. In problems including sparse recovery
\cite{kelner2023semi} and matrix completion \cite{cheng2018non},
methods and analyses that work well for uniform sampling
can fail dramatically with a semi-random adversary. In this paper, we investigate top-$K$ ranking under semi-random sampling. Our goal is to address the following question:

\begin{centering}{\bigskip{}
\emph{Can we identify the top-$K$ items with minimal sample complexity,
even under a monotone adversary?}
\bigskip{}
}\end{centering}

\subsection{Technical challenges}
Top-$K$ ranking under semi-random sampling brings some unique challenges. To begin with, it is worth noting that bounding the $\ell_2$ estimation error for the score vector $\bm{\theta}^\star$ is 
not sufficient to guarantee exact recovery of the top-$K$ items with an optimal sample complexity; 
see~\cite{chen2015spectral, jang2016top, chen2019spectral}. 
Instead, one would  need a more fine-grained $\ell_\infty$ error bound.

To further complicate matters, even under uniform sampling, obtaining optimal control of the $\ell_\infty$ error---thus ensuring optimal sample complexity for top-$K$ ranking---poses a significant challenge. \cite{chen2019spectral} and its follow-up work \cite{chen2022partial,gao2023uncertainty} 
successfully characterize the optimal $\ell_\infty$ error of the MLE, leveraging a powerful leave-one-out argument~\cite{el2018impact,zhong2018near,abbe2020entrywise,ma2018implicit,chen2019gradient}; see~\cite{chen2021spectral} for more references. 
However, a successful 
application of this argument relies crucially on the independence of the edges and certain homogeneity (e.g., degree homogeneity) in the Erd\H{o}s\textendash R\'enyi random graphs---a model for uniform sampling. These properties are easily violated 
for general comparison graphs, let alone the semi-random model we consider herein. 
In fact, as a manifestation, \cite{li2022ell_} recently applies the leave-one-out technique 
to the BTL model in a general (deterministic) sampling mechanism, and obtains a somewhat loose control on the $\ell_\infty$ error of the 
MLE. Notably, even in the special case of uniform sampling, the required sample complexity for MLE can be $n$ times 
larger than the optimal one.  As a result, to address the problem of top-$K$ ranking with a monotone adversary, one needs to develop 
a novel analysis that goes beyond uniform sampling and the leave-one-out technique.

\subsection{Main contributions}
The key result of this paper is to answer the main question affirmatively: 
we show that the weighted maximum likelihood estimator (MLE) with proper choices of weights is 
able to recover the top-$K$ items with near-optimal sample complexity, albeit under a semi-random adversary. 
Moreover, the weights can be computed efficiently, in nearly-linear time in the size of the comparison graph. 
We achieve this through a combination of analytical and algorithmic innovations, which we detail below. 

\paragraph{Analytical contributions.}
We provide a novel $\ell_\infty$ error analysis of the weighted MLE 
with explicit dependence on the spectral properties of the weighted comparison graph 
(e.g., the maximum degree, and the spectral gap of the weighted graph Laplacian); see Theorem~\ref{thm:weight_MLE_spectral}. 
While the dependence on 
spectral properties has been characterized for the $\ell_2$ error of the MLE~\cite{shah2016estimation,hajek2014minimax}, we remark again that the $\ell_2$ error alone cannot guarantee top-$K$ recovery with optimal sample complexity. 

Inspired by the recent work~\cite{chen2023ranking}, we analyze the weighted MLE via a preconditioned gradient descent method that iteratively approximates the weighted MLE. This analysis bypasses the use of the 
leave-one-out argument. 
As opposed to some mysterious and complicated functions for the comparison graph appearing in the performance bound (cf.~Theorem~1 in~\cite{chen2023ranking}), our characterization of the $\ell_\infty$ error of the weighted MLE depends explicitly on the spectral properties of the weighted comparison graph. In particular, it is tight when applied to uniform sampling, in stark contrast with the previously mentioned result in \cite{li2022ell_}.
We expect this novel $\ell_\infty$ error analysis to be broadly applicable to more general sampling mechanisms beyond the semi-random case. 

\paragraph{Algorithmic contributions. }
Motivated by the $\ell_\infty$ analysis of the weighted MLE, our goal boils down to 
finding a reweighting of the semi-random comparison graph such that it  
satisfies the required spectral properties. These amount to a constant lower bound on the spectral gap and upper bounds on the maximum degree and maximum weight in the reweighting.  
Taking a convex optimization approach, we show that the problem of finding such a reweighting can be cast exactly as a semi-definite program (SDP).
We then develop a fast first-order method---based on the Matrix Multiplicative Weight Update (MMWU) framework~\cite{aroraCombinatorialPrimalDualApproach2016} to approximately solve the resulting SDP; see Algorithm~\ref{alg:MMWU}. We further show that such an approximate solution yields a desired set of weights, and the solution can be found efficiently, in nearly-linear time in the size of the semi-random graph.  
We believe that our SDP approach may find broader applications in learning problems over semi-random graphs where we need to restore spectral properties that have been disrupted by adversarial perturbations.




\subsection{Related work}

\paragraph{Ranking with the BTL model.}

The BTL model is a classic model for the ranking problem and has been extensively studied
in the literature. Various methods have been proposed to tackle this problem, including Borda counting \cite{borda1781memoire}, the maximum
likelihood estimator~\cite{ford1957solution}, and the spectral method~\cite{Negahban2012RankCR}, among others. 
Since the conventional $\ell_2$ analysis~\cite{Negahban2012RankCR} fails to capture the accuracy of top-$K$ recovery, recent advances \cite{chen2015spectral,jang2016top,chen2019spectral,chen2022partial} focus on 
establishing the $\ell_{\infty}$ estimation error of the score vector $\bm{\theta}^{\star}$. 
The story is mostly successful under the uniform sampling model. 
For instance, \cite{chen2019spectral} first shows that both the spectral method and the (regularized) MLE achieve minimax optimal $\ell_{\infty}$ estimation error, and recover 
exactly the top-$K$ items under the minimal sample complexity. \cite{chen2022partial} further proves that the vanilla MLE without regularization is optimal, and is superior to the spectral method in terms of the leading constant in the sample complexity. As uniform sampling is often too ideal, 
several attempts have been made to go beyond it.
Most recently, \cite{li2022ell_} and \cite{chen2023ranking} 
investigate the $\ell_{\infty}$ guarantee of the MLE for general comparison graphs. 
As we mentioned,  
their analyses are loose, even when applied to the special case of uniform sampling. 


\paragraph{Semi-random adversary.}

Semi-random adversary has been examined in a number of settings.
Early work in this line studies problems related to semi-random
graphs, including graph partitioning \cite{makarychev2012approximation},
coloring \cite{blum1995coloring,feige2001heuristics}, and finding
independent sets \cite{feige2001heuristics}. In recent years, researchers 
have started to consider semi-random adversary in non-graphical data structures
such as sparse recovery \cite{kelner2023semi}, Gaussian mixture model
\cite{awasthi2018clustering}, matrix sensing \cite{gao2023robust}, and dueling optimization~\cite{blum2023dueling}.
While the exact definition of semi-random varies, it usually involves
some seemingly benign manipulation on top of random sampling.
For instance, \cite{liu2022minimax,moitra2016robust,makarychev2016learning,fei2020achieving}
study stochastic block model, where the data is corrupted by monotone
adversary that can arbitrarily add edges within the clusters and remove
edges between the clusters. \cite{cheng2018non} considers the low-rank
matrix completion problem where the adversary can only provide additional
observed entries. 


\paragraph{Notation.}

For a positive integer $k$, we use the shorthand $[k] \coloneqq \{1,2,\ldots, k\}$.
For any symmetric matrix $\bm{A},\bm{B}\in\mathbb{R}^{n\times n}$,
$\bm{A}\preceq\bm{B}$ means $\bm{B}-\bm{A}$ is positive semidefinite,
i.e., $\bm{v}^{\top}(\bm{B}-\bm{A})\bm{v}\ge0$ for any $\bm{v}\in\mathbb{R}^{n}$.
For any real symmetric matrix $\bm{A}$, we use $\lambda_{1}(\bm{A})\ge\lambda_{2}(\bm{A})\ge\cdots\ge\lambda_{n}(\bm{A})$
to denote its eigenvalues and $\bm{A}^{\dagger}$ its Moore-Penrose
pseudo-inverse. We use $\bm{e}_{i}$ to denote the standard unit vector
with $1$ at $i$-th coordinate and 0 elsewhere. For any two real number $a$ and $b$, $a\vee b$ denotes the maximum of $a, b$ and $a\wedge b$ denotes the minimum of $a, b$. 
Additionally, the standard notation $f(n)=O\left(g(n)\right)$ or $f(n)\lesssim g(n)$ means that there exists a constant $c>0$ such that $\left|f(n)\right|\leq c|g(n)|$;  $f(n)\gtrsim g(n)$ means that there exists a constant $c>0$ such that $|f(n)|\geq c\left|g(n)\right|$. Also, $f(n)\gg g(n)$ means that there exists some large enough constant $c>0$ such that $|f(n)|\geq c\left|g(n)\right|$; $f(n)\ll g(n)$ means that there exists some sufficiently small constant $c>0$ such that $|f(n)|\leq c\left|g(n)\right|$. 

\section{Main results}
We begin with formally introducing the problem setup for top-$K$ ranking with a monotone adversary. 

\subsection{Problem formulation}
\paragraph{Semi-random comparison graph.}
Let $\mathcal{G}_{\mathrm{}}=(\mathcal{V},\mathcal{E})$ be a comparison
graph over the $n$ items of interest. In other words, items $i$
and $j$ are compared if and only if $(i,j)\in\mathcal{E}$. Prior
work often assumes a homogeneous random comparison graph, e.g., $\mathcal{G}$
is an Erd\H{o}s\textendash R\'enyi random graph where each pair 
$(i,j)$ is an edge with probability~$p$ independently. Our focus in this work 
is to investigate the ranking problem with a monotone adversary described
as follows. Let $\mathcal{G}_{\mathrm{ER}}=(\mathcal{V},\mathcal{E}_{\mathrm{ER}})$
be the initial Erd\H{o}s\textendash R\'enyi random graph. An adversary
observes $\mathcal{G}_{\mathrm{ER}}$ and is allowed to add edges
to $\mathcal{E}_{\mathrm{ER}}$ \emph{arbitrarily}. We denote the semi-random comparison
graph with added edges $\mathcal{G}_{\mathrm{SR}}=(\mathcal{V},\mathcal{E}_{\mathrm{SR}})$.
From now on, $\mathcal{G}_{\mathrm{SR}}$ will be the comparison graph on which pairwise comparisons are made. 

\paragraph{Latent scores and pairwise comparisons.}

In the Bradley-Terry-Luce (BTL) model, each item $i\in[n]$ is associated
with a latent score $\theta_{i}^{\star}$ that represents the skill
level of item $i$. Without loss of generality, we assume that the
scores are ordered, i.e., $\theta_{1}^{\star}\ge\theta_{2}^{\star}\geq\cdots\geq\theta_{n}^{\star}$. 

For each pair $(i,j)\in\mathcal{E}_{\mathrm{SR}}$ with $i>j$, we observe $L$
outcomes $\{y_{ij}^{(l)}\}_{l\in[L]}$, which are independent Bernoulli
random variables obeying 
\[
\mathbb{P} ( y_{ij}^{(l)}=1 ) =\frac{e^{\theta_{j}^{\star}}}{e^{\theta_{i}^{\star}}+e^{\theta_{j}^{\star}}}.
\]
Correspondingly, we define the average winning rate $y_{ij}\coloneqq(1/L)\sum_{l=1}^{L}y_{ij}^{(l)}$. 
A simple observation is that the BTL model is shift invariant in $\bm{\theta}^\star$ so we assume $\bm{1}_{n}^{\top}\bm{\theta}^\star=0$ without loss of generality. 
Finally, we define a sort of condition number $\kappa\coloneqq e^{\theta_{1}^\star}/e^{\theta_{n}^\star}$
to characterize the range of $\bm{\theta}^\star$.

\paragraph{Top-$K$ recovery. }
Our goal is to recover the top-$K$ items. Clearly, the hardness of the problem relies on the score 
difference between the $K$-th and the $(K+1)$-th preferred items, which we denote by
\begin{equation*}
    \Delta_{K} \coloneqq \theta^\star_{K} - \theta^\star_{K+1}.
\end{equation*}
Throughout the paper, we assume $\Delta_{K} > 0$ so that the top-$K$ items are well-defined, and are given by $[K]$.

Now we turn to the main message of this paper: the weighted MLE, with proper weights, achieves exact top-$K$ recovery. 

\subsection{Weighted MLE achieves exact recovery}

\begin{algorithm}[t]
\caption{\label{alg:main}Weighted MLE for top-$K$ recovery under the semi-random
sampling}

\begin{enumerate}
\item Observe $\mathcal{G}_{\mathrm{SR}}=(\mathcal{V},\mathcal{E}_{\mathrm{SR}})$
and $\{y_{ij}\}_{(i,j)\in\mathcal{E}_{\mathrm{SR}},i>j}$.
\item Use Algorithm~\ref{alg:MMWU} with appropriate
input to compute $\{w_{ij}\}.$ The input does not include $\{y_{ij}\}_{(i,j)\in\mathcal{E}_{\mathrm{SR}},i>j}$.
\item Output $\widehat{\bm{\theta}}\coloneqq\arg\min_{\bm{\theta}:\bm{1}_{n}^{\top}\bm{\theta}=0}\mathcal{L}_{w}(\bm{\theta})$.
\end{enumerate}
\end{algorithm}


Under uniform sampling, it has been shown in the work~\cite{chen2019spectral}
that the MLE achieves exact recovery
of the top-$K$ items with optimal sample complexity. This motivates
us to consider a weighted MLE for the semi-random graph $\mathcal{G}_{\mathrm{SR}}$ that can approximate the vanilla MLE under the purely random sampling case (i.e., with the comparison graph being $\mathcal{G}_{\mathrm{ER}}$). 
More formally, let $\{w_{ij}\}_{i>j}$ be a set of non-negative
weights supported on $\mathcal{E}_{\mathrm{SR}}$, that is, $w_{ij}=0$
if $(i,j)\notin\mathcal{E}_{\mathrm{SR}}$. We define the weighted
negative log-likelihood function $\mathcal{L}_{\bm{w}}(\cdot)$:
\begin{align}
\mathcal{L}_{w}(\bm{\theta}) & \coloneqq-\frac{1}{L}\sum_{i,j:i>j}\sum_{l=1}^{L}w_{ij}\log\left(y_{ji}^{(l)}\frac{e^{\theta_{i}}}{e^{\theta_{i}}+e^{\theta_{j}}}+(1-y_{ji}^{(l)})\frac{e^{\theta_{j}}}{e^{\theta_{i}}+e^{\theta_{j}}}\right)\nonumber\\
 & =\sum_{i,j:i>j}w_{ij}\left(-y_{ji}(\theta_{i}-\theta_{j})+\log(1+e^{\theta_{i}-\theta_{j}})\right)\label{eq:MLE_loss}, 
\end{align}
where we recall $y_{ji}=(1/L)\sum_{l=1}^{L}y_{ji}^{(l)}$. We then
define the weighted MLE to be 
\begin{equation}
\widehat{\bm{\theta}}\coloneqq\arg\min_{\bm{\theta}:\bm{1}_{n}^{\top}\bm{\theta}=0}\quad\mathcal{L}_{w}(\bm{\theta}).\label{eq:weighted_MLE}
\end{equation}
The top-$K$ items are identified by selecting the top-$K$ entries
in the estimate $\widehat{\bm{\theta}}$. 

The key to the success of the weighted MLE lies in a proper construction of the set of weights $\{w_{ij}\}$ that
can mimic the vanilla MLE under an Erd\H{o}s\textendash R\'enyi random graph.
It turns out that such a goal can be achieved, and we have the following
guarantees for the estimation error as well as the top-$K$ recovery
performance. 


\begin{theorem}\label{thm:weighted_MLE_semirandom}

Suppose that $np\ge C_{1}\log (n)$ and $npL\ge C_{2}\kappa^{4}\log^3(n)$
for some large enough constants $C_{1},C_{2}>0$. With probability
at least $1-n^{-10}$, Algorithm~\ref{alg:main} returns the weighted MLE $\widehat{\bm{\theta}}$
that obeys
\begin{equation}
\|\widehat{\bm{\theta}}-\bm{\theta}^{\star}\|_{\infty}\le C_{3}\kappa\sqrt{\frac{\log(n)}{npL}}\label{eq:l_infty_guarantee_semirandom}
\end{equation}
for some constant $C_{3}>0$. On this event, the top-$K$ items are
recovered exactly as long as 
\[
n^2 pL\ge C_{4}\frac{\kappa^{2}n \log(n)}{\Delta_{K}^{2}}
\]
for some large enough constant $C_{4} > 0$.
In addition, the reweighting algorithm (i.e., Algorithm~\ref{alg:MMWU}) runs in nearly-linear time in the 
size of $\mathcal{G}_{\mathrm{SR}}$.
\end{theorem} 

We defer the details on the construction of the weights to Section~\ref{sec:reweighting} and focus on 
interpreting the performance of the weighted MLE now. Similar to the literature~\cite{chen2015spectral,chen2019spectral,chen2022partial}, we 
assume $\kappa = O(1)$ when interpreting  the results. 

\paragraph{Near-optimal sample complexity under monotone adversary.}
Under the uniform sampling, the minimax sample complexity for top-$K$ recovery has been identified in~\cite{chen2015spectral}. 

\begin{figure}
    \centering
    \includegraphics[width = 0.45\textwidth]{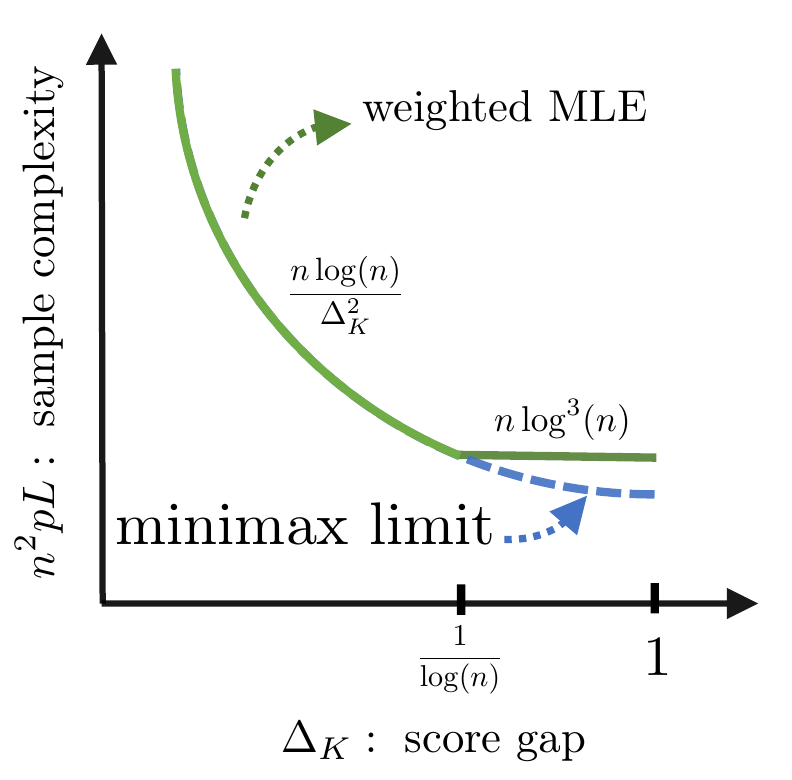}
    \caption{Sample complexity required to exactly recover top-$K$ items v.s.~score gap $\Delta_K$. The solid line represents the required sample complexity of the weighted MLE as given in Theorem~\ref{thm:weighted_MLE_semirandom}. The dashed line represents the minimax lower bound given in Theorem~\ref{thm:minimax}.}
    \label{fig:sample_complexity}
\end{figure}

\begin{theorem}[Informal]\label{thm:minimax}
    Assume that $\kappa=O(1)$. 
    If $n^2 pL \leq c n \log(n) / \Delta_{K}^2$ for some small constant $c>0$, then no method whatsoever can achieve exact recovery with constant probability. 
\end{theorem}
Since uniform sampling is a special case of semi-random sampling (i.e., the adversary adds no edges at all), 
Theorem~\ref{thm:minimax} is also a valid lower bound for the semi-random case. 
Comparing the performance of the weighted MLE (Theorem~\ref{thm:weighted_MLE_semirandom}) and the lower bound (Theorem~\ref{thm:minimax}), 
we see that the weighted MLE achieves near-optimal sample complexity for top-$K$ recovery with a monotone adversary; 
see Figure~\ref{fig:sample_complexity}.
More precisely, when $\Delta_{K} \lesssim 1 / \log(n)$, the weighted MLE requires $n^2 p L \asymp n \log(n) / \Delta_{K}^2$ 
number of observations, which is exactly the minimax limit. 
On the other hand, when $1 / \log(n) \lesssim \Delta_{K} \lesssim 1$, the weighted MLE needs 
$n^2 p L \asymp n \log^3(n)$ comparisons, which exceeds the minimax lower bound by at most a $\log^2(n)$ factor.

\paragraph{Computational complexity.} Compared to a vanilla MLE, Algorithm~\ref{alg:main} requires an additional reweighting step, which is performed by Algorithm~\ref{alg:MMWU}. Crucially, we show that thsi step does not fundamentally alter the complexity of the whole procedure, as Algorithm~\ref{alg:MMWU}  runs in nearly-linear time in the size of the graph $\mathcal{G}_{\mathrm{SR}}$. 
To ensure this fast running time, we combine the  Matrix Multiplicative Weight Update (MMWU) framework~\cite{aroraCombinatorialPrimalDualApproach2016} with a number of known approximation schemes, including fast computations of the action of the matrix exponential~\cite{orecchiaApproximatingExponentialLanczos2012}, randomized dimensional reduction~\cite{achlioptasDatabasefriendlyRandomProjections2003} and fast solvers for packing linear programs~\cite{packing}. We also describe a simpler algorithm which does not require solving a packing linear program as a subroutine but relies on an easy-to-compute greedy approximation. This is the algorithm we implement in Section~\ref{app:experiments}.

\paragraph{Improved dependence on $\kappa$.} 
We remark in passing that compared to the $\ell_\infty$ bound $\|\widehat{\bm{\theta}}-\bm{\theta}^{\star}\|_{\infty}\lesssim \kappa^2 \sqrt{\frac{\log(n)}{npL}}$ provided in~\cite{chen2019spectral}, which is based on a probabilistic leave-one-out argument, our $\ell_\infty$ error bound~\eqref{eq:l_infty_guarantee_semirandom} is tighter by a $\kappa$ factor. 



\section{Two algorithmic components}

In this section we present the high-level analyses of the two algorithmic components
in Algorithm~\ref{alg:main}, namely the weighted MLE, and the SDP-based reweighting method. 

\subsection{Weighted MLE}\label{sec:MLE}
As the selected weight can be dependent on all edges of the comparison
graph, we cannot use the popular and powerful leave-one-out technique
in recent papers \cite{chen2019spectral,chen2022partial} to achieve
entrywise control. Instead, we rely on and refine a new analysis for MLE~\cite{chen2023ranking} 
that is geared towards general graphs.

Recall that our estimator is a weighted MLE with weights $\{w_{ij}\}$ 
supported on the edges of the semi-random graph $\mathcal{G}_{\mathrm{SR}}$. 
Let $w_{\max} \coloneqq \max_{i,j}w_{ij}$ be the maximum weight, $d_{\max} \coloneqq \max_{i\in[m]}\sum_{j:j\neq i}w_{ij}$ be the maximum (weighted) degree, and $d_{\min} \coloneqq \min_{i\in[m]}\sum_{j:j\neq i}w_{ij}$ be the minimum (weighted) degree. 
In addition, we define the weighted graph Laplacian of $\mathcal{G}_{\mathrm{SR}}=(\mathcal{V},\mathcal{E}_{\mathrm{SR}}, \{w_{ij}\})$ to be
\begin{equation}
    \bm{L}_{w} \coloneqq \sum_{(i,j):i>j}w_{ij}(\bm{e}_{i}-\bm{e}_{j})(\bm{e}_{i}-\bm{e}_{j})^{\top}.\label{eq:L_w}
\end{equation}



We have the following performance bounds for the weighted MLE whenever the weights are 
independent with the observed comparisons $\{y_{ij}\}$. 
\begin{theorem}
\label{thm:weight_MLE_spectral}
Suppose that the weighted graph $\mathcal{G}_{\mathrm{SR}}=(\mathcal{V},\mathcal{E}_{\mathrm{SR}}, \{w_{ij}\})$ is connected. Assume that $w_{\max}\le n^2$ and $d_{\min}\ge 1$.
Further suppose that
\[L\ge C_1\frac{\kappa^4 w_{\max} (d_{\max})^4\log^2(n)}{(\lambda_{n-1}(\bm{L}_{w}))^5} \label{eq:L}\]
for some large enough constant $C_1 > 0$. Then with probability at least $1-n^{-10}$, we
have 
\[
\|\widehat{\bm{\theta}}-\bm{\theta}^\star\|_\infty\le C_2 \kappa\sqrt{\frac{ w_{\max} \log(n)}{\lambda_{n-1}(\bm{L}_{w})L}},
\]
 where $C_2>0$ is a constant. On this event, the top-$K$ items are
recovered exactly as long as 
\[
L\ge C_{3}\frac{\kappa^{2} w_{\max} \log(n)}{\lambda_{n-1}(\bm{L}_w)\Delta_{K}^{2} }
\]
for some large enough constant $C_{3} > 0$.

\end{theorem}
\noindent See Section~\ref{subsec:Analysis_wMLE} for the proof of this theorem. Note that the assumptions on $w_{\max}$  and $d_{\min}$ are mild and stated only to simplify the log factor.  
\medskip

Theorem~\ref{thm:weight_MLE_spectral} provides an $\ell_{\infty}$ error bound for the weighted MLE under general comparison
graphs. The bounds depend explicitly on the spectral properties of the weighted graph, including the maximum degree, and the spectral gap of the graph Laplacian. 

An important by-product of our novel analysis of the weighted MLE is to demonstrate the 
optimal $\ell_{\infty}$ estimation error of the vanilla MLE in the uniform sampling case. 

\begin{corollary}\label{coro:ER}
    Consider the uniform sampling case, that is the comparison graph is an Erd\H{o}s\textendash R\'enyi graph. 
    Assume that $\kappa=O(1)$, and that $p \gtrsim \log(n) / n$. The vanilla MLE with high probability achieves 
    \[
\|\widehat{\bm{\theta}}-\bm{\theta}^\star\|_\infty\lesssim \sqrt{\frac{\log(n)}{npL}}, 
\]
as long as $npL \gtrsim \log^2(n)$.
\end{corollary}

\begin{proof}
    Observe that the vanilla MLE is equivalent to the weighted MLE with a uniform weight $1$ on all the edges. 
    With this choice, it is easy to show (see Lemma~\ref{lemma:spectral_ER})  that with high probability, 
    \begin{subequations}\label{subeq:nice-spectral}
        \begin{align}
            w_{\max} &\le  1,\label{eq:rw-lemma-max}\\
            d_{\max} &\le 2np,  \label{eq:rw-lemma-degree}\\
            \lambda_{n-1}(\bm{L}_{w}) &\ge Cnp, \label{eq:rw-lemma-spectral-gap}
    \end{align}
    \end{subequations}
where $C>0$ is some constant. Moreover by Lemma~\ref{lemma:d_min}, $\lambda_{n-1}(\bm{L}_{w}) \ge np/2$ together with~$p\gtrsim \log(n)/ n$ implies $d_{\min}\ge  1$.  Apply Theorem~\ref{thm:weight_MLE_spectral} verbatim to arrive at the desired conclusion. 
\end{proof}

\subsection{An SDP-based reweighting}\label{sec:reweighting}

In view of the proof of Corollary~\ref{coro:ER}, to mimic the vanilla MLE under uniform sampling, 
it is sufficient to construct the weights $\{w_{ij}\}$ such that 
the weighted graph $\mathcal{G}_{\mathrm{SR}}=(\mathcal{V},\mathcal{E}_{\mathrm{SR}}, \{w_{ij}\})$ satisfies 
the spectral properties~\eqref{subeq:nice-spectral}.
In this section, we describe how to formulate this problem as a saddle-point semi-definite program (SDP) and approximately solve it in nearly-linear time in the size of $\cG_{\mathrm{SR}}$ by designing a fast {first-order} method.
%

We formulate our task in terms of a convex optimization problem with variables $\{w_{ij} \in \R\}_{(i,j) \in \cE_{\mathrm{SR}}}$ representing our desired reweighting. The convex feasible set $\cF$ for such weights is given by the rescaled Equations~\eqref{eq:rw-lemma-degree} and \eqref{eq:rw-lemma-spectral-gap}:
\begin{equation}\label{eq:w-feasible-set}
\cF \defeq \{\bm{w} \in \R^{\cE_{\mathrm{SR}}}_{\geq 0} : \forall i \in \mathcal{V}, \sum_{j:(i,j) \in \cE_{\mathrm{SR}}} w_{ij} \leq 2np \; \land \; \forall (i,j) \in \cE_{\mathrm{SR}}, w_{ij} \leq 1 \}.   
\end{equation}
For a choice of weights $\bm{w}$, we denote by $\bm{L}_{\bm{w}}$ the corresponding weighted Laplacian as defined in (\ref{eq:L_w}).
With this notation, we consider the problem of maximizing the spectral gap $\lambda_{n-1}(\bm{L}_{\bm{w}})$  over $\bm{w} \in \cF$. 
It is a well-known fact that this is a convex optimization problem in the variables $\bm{w}$~\cite{boydFastestMixingMarkov2004}. Indeed, we can write $\lambda_{n-1}(\bm{L}_{\bm{w}})$ as the minimum of the matrix inner product $\langle \bm{L}_{\bm{w}}, \bm{X}\rangle$ over $\bm{X}$ in the set 
$$
\Delta \defeq \{\bm{X} \in \mathbb{S}^\mathcal{V}: \bm{X} \succeq 0 \; \land \; \langle \Pi_{\perp \bm{1}}, \bm{X} \rangle = 1\},
$$
where $\mathbb{S}^\mathcal{V}$ is the set of symmetric linear operators over $\mathcal{V}$ and $\Pi_{\perp \bm{1}}$ is the orthogonal projector over the orthogonal complement of the vector $\bm{1}$, which is the eigenvector of $\bm{L}_{\bm{w}}$ with the smallest eigenvalue.
Therefore, our desired convex optimization problem can be recast as the following saddle point problem between an SDP variable $\bm{X}$ and the weight $\bm{w}$. This formulation will be crucial for the solvers designed in the later section:
\begin{equation}\label{eq:sdp-saddle-point}
\mathrm{OPT} \coloneqq  \max_{\bm{w} \in \cF} \min_{\bm{X} \in \Delta} \quad \langle \bm{L}_{\bm{w}} , \bm{X}\rangle. \tag{Saddle-Point SDP}
\end{equation}

By considering the weighting $\bm{w}$ corresponding to the original graph in $\cG_{\mathrm{ER}}$ as a feasible reweighting for~\ref{eq:sdp-saddle-point}, the proof of Corollary~\ref{coro:ER} immediately implies a lower bound on $\mathrm{OPT}$: with probability at least $1-n^{-10}$,  
\begin{equation}
    \mathrm{OPT} \geq \frac {np}{2}, \label{eq:OPT}
\end{equation}
as long as $p \geq C\log(n) / n$ for some sufficiently large constant $C>0$.

While it is not always possible to recover the underlying graph $\cG_{\mathrm{ER}}$, the lower bound~\eqref{eq:OPT} ensures that, by approximately solving~\ref{eq:sdp-saddle-point}, we can find a reweighting of $\cG_{\mathrm{SR}}$ that satisfies the required spectral properties~\eqref{subeq:nice-spectral}. The next lemma, proved in Section~\ref{sec:sdp}, shows that this approximate solution can be computed in nearly-linear time in the size of $\cG_{\mathrm{SR}}$. The corresponding algorithm, Algorithm~\ref{alg:MMWU} to be detailed in Section~\ref{sec:sdp}, is based on the Matrix Multiplicative Weight Update (MMWU) framework~\cite{aroraCombinatorialPrimalDualApproach2016}, a first-order method for non-smooth convex SDP optimization.


\begin{lemma}\label{lemma:reweighting} 
Given the observed comparison
graph $\mathcal{G}_{\mathrm{SR}}$, there is an algorithm that computes
a set of non-negative weights $\left\{ w_{ij}\right\} _{i>j}$ supported
on $\mathcal{E}_{\mathrm{SR}}$ that satisfy the properties~\eqref{subeq:nice-spectral} with high probability.
In addition, the algorithm runs in nearly-linear time in the size of $\mathcal{G}_{\mathrm{SR}}.$
      

\end{lemma} 

We finish this section by noting that once the spectral properties are met, repeating the proof of Corollary~\ref{coro:ER} yields the desired results in Theorem~\ref{thm:weighted_MLE_semirandom} in the semi-random case. 



%




\section{Analysis of weighted MLE \label{subsec:Analysis_wMLE}}
In this section, we present detailed analysis of the weighted MLE with the aim of proving Theorem~\ref{thm:weight_MLE_spectral}.

Given a set of weights $\{w_{ij}\}$, the Hessian of the weighted loss at ground
truth is given by
\[
\nabla^{2}\mathcal{L}(\bm{\theta}^{\star})=\sum_{(i,j):i>j}w_{ij}\underbrace{\frac{e^{\theta_{i}^{\star}}e^{\theta_{j}^{\star}}}{(e^{\theta_{i}^{\star}}+e^{\theta_{j}^{\star}})^{2}}}_{\eqqcolon z_{ij}}(\bm{e}_{i}-\bm{e}_{j})(\bm{e}_{i}-\bm{e}_{j})^{\top}.
\]
This is exactly the graph Laplacian of the weighted graph $\mathcal{G}_{\mathrm{SR}}$
with weights $\{w_{ij}z_{ij}\}$. Hence we denote this by $\bm{L}_{wz}$. We also define
the effective resistance to be 
\[
\bm{\Omega}_{kl}(\bm{L}_{wz})\coloneqq(\bm{e}_{k}-\bm{e}_{l})^{\top}\bm{L}_{wz}^{\dagger}(\bm{e}_{k}-\bm{e}_{l}),
\]
where $\bm{L}_{wz}^{\dagger}$ is the pseudo-inverse of $\bm{L}_{wz}$.

Inspired by Theorem~1 in the paper~\cite{chen2023ranking}, the first step of the proof 
relates the performance of the weighted MLE with two crucial quantities $\{B_{kl}\}$ and $\{Q_{kl}\}$.
\begin{lemma}\label{thm:weighted_MLE}

Suppose that the weighted graph $(\mathcal{V},\mathcal{E}_{\mathrm{SR}}, \{w_{ij}\})$ is connected by edges of non-zero weight 
and the weights $\{w_{ij}\}$ are independent with the observations
$\{y_{ij}\}$. For any $(k,l)\in[n]^2, k\neq l$, let $B_{kl}$ and $Q_{kl}$ be
some quantities obeying
\begin{subequations}
   \begin{align}
B_{kl} & \ge C\sqrt{\frac{\kappa}{L}\left(\max_{i,j}w_{ij}\right)\bm{\Omega}_{kl}(\bm{L}_{wz})\log(n)};\label{eq:B_def}\\
Q_{kl} & \ge\sum_{(i,j):i>j}w_{ij}B_{ij}^{2}\left|(\bm{e}_{k}-\bm{e}_{l})^{\top}\bm{L}_{wz}^{\dagger}(\bm{e}_{i}-\bm{e}_{j})\right|.\label{eq:Q_def}
\end{align} 
\end{subequations}
Here $C>0$ is some large enough constant. Suppose that $Q_{kl}\le4B_{kl}$
for any $(k,l)$. Then with probability at least $1-n^{-10}$, we
have that for any $(k,l)\in[n]^{2}$, 
\[
\|\widehat{\bm{\theta}} - \bm{\theta}^\star \|_\infty \le B_{kl}.
\]On this event, the top-$K$ items are
recovered exactly as long as 
$$\max_{k, l} B_{kl}\le \Delta_K/2.$$
\end{lemma}
Admittedly, the two quantities $\{B_{kl}\}$ and $\{Q_{kl}\}$ appearing in the performance 
bound of MLE is quite mysterious. 
A key contribution of this paper is to further relate these two quantities to basic spectral properties of the 
weighted graph $(\mathcal{V},\mathcal{E}_{\mathrm{SR}}, \{w_{ij}\})$. 
We start with the characterization of $B_{kl}$. Recall that $w_{\max} = \max_{i,j}w_{ij}$ is the maximum weight. 
\begin{lemma}\label{lem:B}
For any $(k,l)\in [n]^2$, the effective resistance $\Omega_{kl}(\bm{L}_{wz})$ satisfies   \[\Omega_{kl}(\bm{L}_{wz})\le \frac{8\kappa}{\lambda_{n-1}(\bm{L}_{w})}.\]
As a result, it is sufficient to take
\[
B_{kl} = C \kappa \sqrt{\frac{w_{\max} \log(n)}{L\lambda_{n-1}(\bm{L}_w)}},
\]where $C>0$  is some large enough constant.

\end{lemma}
Now we move on to controlling the $Q$ factors via spectral properties of the weighted graph. 
Recall $d_{\max} = \max_{i\in[n]}\sum_{j:j\neq i}w_{ij}$ is the maximum (weighted) degree and $d_{\min} = \min_{i\in[n]}\sum_{j:j\neq i}w_{ij}$ is the minimum (weighted) degree. 
We have the following bound.
\begin{lemma}\label{lem:Q}
Suppose that $w_{\max} \le n^2$ and that $d_{\min}\ge 1$. Then for any $(k,l)\in [n]^2$, we have
\[\sum_{(i,j):i>j} w_{ij}\left|(\bm{e}_{k}-\bm{e}_{l})^{\top}\bm{L}_{wz}^{\dagger}(\bm{e}_{i}-\bm{e}_{j})\right|\le C\kappa \cdot\frac{(d_{\max})^2 \log(n)  }{(\lambda_{n-1}(\bm{L}_w))^2},\]
where $C_1>0$ is some constant. As a result, it is sufficient to take 
\[Q_{kl} = C_2\kappa^3 \cdot\frac{w_{\max}(d_{\max})^2\log^2(n) }{L (\lambda_{n-1}(\bm{L}_w))^3}\]for some large enough constant $C_2>0$.
\end{lemma}

Combining Lemmas~\ref{lem:B}-\ref{lem:Q}, we see that $Q_{kl}\le 4 B_{kl}$ holds as long as
\[L\ge C_3\frac{\kappa^4 w_{\max} (d_{\max})^4 \log^3(n)}{(\lambda_{n-1}(\bm{L}_{w}))^5}\] for some large enough constant $C_3>0$. This together with Lemma~\ref{thm:weighted_MLE} 
completes the proof of Theorem~\ref{thm:weight_MLE_spectral}. In what follows, we present the proofs of Lemmas~\ref{lem:B}-\ref{lem:Q}, and defer the proof of Lemma~\ref{thm:weighted_MLE} to Appendix~\ref{sec:proof-weighted-MLE-B-Q}. 


\subsection{Proof of Lemma~\ref{lem:B}}


Recall that $\bm{\Omega}_{kl}(\bm{L}_{wz})=(\bm{e}_{k}-\bm{e}_{l})^{\top}\bm{L}_{wz}^{\dagger}(\bm{e}_{k}-\bm{e}_{l})$, which implies  
\[
\bm{\Omega}_{kl}(\bm{L}_{wz})\leq2\|\bm{L}_{wz}^{\dagger}\|=\frac{2}{\lambda_{n-1}(\bm{L}_{wz})}.
\]
Regarding $\lambda_{n-1}(\bm{L}_{wz})$, by definition, one has
\begin{align}
\lambda_{n-1}(\bm{L}_{wz}) & =\inf_{\bm{v}\in\mathbb{R},\|\bm{v}\|=1,\bm{v}\perp\bm{1}_{n}}\bm{v}^{\top}\bm{L}_{wz}\bm{v}\nonumber \\
 & =\inf_{\bm{v}\in\mathbb{R},\|\bm{v}\|=1,\bm{v}\perp\bm{1}_{n}}\bm{v}^{\top}\left[\sum_{(i,j):i>j}w_{ij}z_{ij}(\bm{e}_{i}-\bm{e}_{j})(\bm{e}_{i}-\bm{e}_{j})^{\top}\right]\bm{v}\nonumber \\
 & \ge\frac{1}{4\kappa}\inf_{\bm{v}\in\mathbb{R},\|\bm{v}\|=1,\bm{v}\perp\bm{1}_{n}}\bm{v}^{\top}\left[\sum_{(i,j):i>j}w_{ij}(\bm{e}_{i}-\bm{e}_{j})(\bm{e}_{i}-\bm{e}_{j})^{\top}\right]\bm{v}\nonumber \\
 & =\frac{1}{4\kappa}\lambda_{n-1}(\bm{L}_{w}),\label{eq:L_wz_to_L_z_eigen}
\end{align}
where the inequality follows from Lemma~\ref{lemma:z_range}.

\subsection{Proof of Lemma~\ref{lem:Q}}

For readers' convenience, we copy the key quantity appearing in Lemma~\ref{lem:Q} below: 
\begin{equation}
\sum_{(i,j):i>j}w_{ij}\left|(\bm{e}_{k}-\bm{e}_{l})^{\top}\bm{L}_{wz}^{\dagger}(\bm{e}_{i}-\bm{e}_{j})\right|.\label{eq:congestion}
\end{equation}
It turns out that this quantity  is closely related to the so-called
conductance of the weighted graph defined as follows.

\begin{definition}[Conductance]\label{def:conductance} For a weighted graph $\mathcal{G}=(\mathcal{V},\mathcal{E}, w)$, we define its conductance to be
\[
\Phi(\mathcal{G})\coloneqq\min_{\mathcal{S}\subset\mathcal{V}}\frac{\sum_{i\in\mathcal{S},j\in\mathcal{V}\setminus\mathcal{S}}w_{ij}}{\min\{\mathrm{vol}(\mathcal{S}),\mathrm{vol}(\mathcal{V}\setminus\mathcal{S})\}},
\]
where the volume of a vertex set $\mathcal{S} \subset \mathcal{V}$ is defined by
\[
\mathrm{vol}(\mathcal{S})\coloneqq\sum_{i\in\mathcal{S}}\sum_{j\in\mathcal{V}:j\neq i}w_{ij}.
\]

\end{definition}

The following lemma links the quantity in (\ref{eq:congestion}) with
graph conductance. This lemma is modified from Lemma~28 in \cite{kelner2014almost}.
The proof is deferred to Section~\ref{subsec:Proof_cr_conductance}. 

\begin{lemma}\label{lemma:cr_conductance} Consider a graph $\mathcal{G}=(\mathcal{V},\mathcal{E})$
equipped with two set of weights $\{w_{ij}\}$ and $\{\widetilde{w}_{ij}\}$
both supported on $\mathcal{E}$. Suppose that the minimum $\{w_{ij}\}$-weighted degree is at least 1. Then one has

\begin{equation}
\sum_{(i,j):i>j}w_{ij}\left|(\bm{e}_{i}-\bm{e}_{j})^{\top}\bm{L}_{\mathcal{G}_{\widetilde{w}}}^{\dagger}(\bm{e}_{k}-\bm{e}_{l})\right|\le\max_{(i,j)\in\mathcal{E}}\left\{ \frac{w_{ij}}{\widetilde{w}_{ij}}\right\} \cdot\frac{8\log\left(\sum_{(i,j):i>j}w_{ij}\right)}{\Phi(\mathcal{G}_{w})^{2}}. \label{eq:congestion_conductance}
\end{equation}

\end{lemma}

Set $\tilde{w}_{ij} = w_{ij} z_{ij}$. We can apply
Lemma~\ref{lemma:cr_conductance} to obtain 

\begin{align*}
\sum_{(i,j):i>j}w_{ij}\left|(\bm{e}_{i}-\bm{e}_{j})^{\top}\bm{L}_{wz}^{\dagger}(\bm{e}_{k}-\bm{e}_{l})\right| & \le\max_{(i,j):w_{ij}>0}\left\{ \frac{w_{ij}}{w_{ij}z_{ij}}\right\} \cdot \frac{8\log\left(\sum_{(i,j):i>j}w_{ij}\right)}{\Phi(\mathcal{G}_{w})^{2}}\\
 & \le4\kappa\cdot\frac{8\log\left(\sum_{(i,j):i>j}w_{ij}\right)}{\Phi(\mathcal{G}_{w})^{2}},
\end{align*}
where the last inequality again follows from Lemma~\ref{lemma:z_range}. 

For the numerator, since $w_{\max} \leq n^2$, we have $\log(\sum_{(i,j):i>j}w_{ij})\le 3\log (n)$.
Now we focus on the denominator, i.e., the graph conductance. 
It is well known that the graph conductance is controlled by the eigenvalue of the normalized Laplacian (see~Lemma~\ref{lemma:conductance_eigen}), that is 
\begin{align*}
    \Phi(\mathcal{G}_{w}) \geq \frac{1}{2} \lambda_{n-1}\left(\bm{D}_{w}^{-1/2}\bm{L}_{w}\bm{D}_{w}^{-1/2}\right),
\end{align*}
where $\bm{D}_{w}$ is a diagonal matrix composed of the weighted degrees. 
By Sylvester's law of inertia (Lemma~\ref{lemma:sylvester}), we further have
\begin{align*}
\lambda_{n-1}\left(\bm{D}_{w}^{-1/2}\bm{L}_{w}\bm{D}_{w}^{-1/2}\right) & \ge\lambda_{n}\left(\bm{D}_{w}^{-1}\right)\lambda_{n-1}\left(\bm{L}_{w}\right) =  (d_{\max})^{-1}\cdot\lambda_{n-1}\left(\bm{L}_{w}\right).
\end{align*}
Taking the above bounds collectively yields the desired claim in Lemma~\ref{lem:Q}.

\renewcommand{\algorithmicrequire}{\textbf{Input:}}
\renewcommand{\algorithmicensure}{\textbf{Output:}}
 \renewcommand{\algorithmiccomment}[1]{// #1}
 
\begin{algorithm}[t] 
\caption{MMWU algorithm for \ref{eq:sdp-saddle-point}}\label{alg:MMWU}
\begin{algorithmic}[1]
\REQUIRE Graph $\cG_{\mathrm{SR}}= (V, \cE_{\mathrm{SR}})$, $|V|=n$, $|\cE_{\mathrm{SR}}|=m,$ error parameter $\epsilon \in (0,\nicefrac{1}{2}]$. 
\STATE $\bm{w}^{(0)} = \bm{0}, \eta=\nicefrac{\epsilon}{4pn}, k = O\left(\nicefrac{\log(n)}{\epsilon^2}\right),$ $T = \nicefrac{8 \log(n)}{\epsilon^2}$
\FOR{$t=0,\ldots,T-1$}
\STATE Sample $\bm{R} = \{\pm \nicefrac{1}{\sqrt{k}} \}^{n \times k}$ uniformly at random. \hfill
\COMMENT{$\{\pm 1\}$ JL matrix with $\nicefrac{1}{\sqrt{k}}$ scaling~\cite{achlioptasDatabasefriendlyRandomProjections2003}} \label{line:jl-matrix}
\STATE Compute 
$\bm{U}^{(t)} = 
\exp\{-\eta \sum_{ij \in \cE_{\mathrm{SR}}} {w}^{(t)}_{ij} \bm{L}_{ij}\} \bm{R}.$ \hfill \COMMENT{Use matrix exponentiation in~\cite{aroraCombinatorialPrimalDualApproach2016}} \label{line:jl-comp}
\STATE Compute $\bm{V}^{(t)} = \nicefrac{\bm{U}^{(t)}}{\sqrt{\langle \Pi_{\perp \bm{1}}, \bm{U}^{(t)} (\bm{U}^{(t)})^T\rangle}} $.\hfill\COMMENT{MMWU Normalization}
\STATE For each $ij \in \cE_{\mathrm{SR}}$, let 
$c_{ij} = \langle \bm{L}_{ij},\bm{V}^{(t)} (\bm{V}^{(t)})^T \rangle$.
%
\hfill\COMMENT{Compute edge gains} \label{line:gains}
\STATE Compute $\hat{\bm{w}}^{(t)} \in \cF$, an $\epsilon$- or $\nicefrac{1}{2}$-approximate solution to $\max_{\bm{w} \in \cF} \langle \bm{c}, \bm{w} \rangle.$ \hfill\COMMENT{Use Theorem~\ref{thm:LP-oracle} or~\ref{thm:greedy-oracle}}
\label{line:oracle}
\STATE Update $\bm{w}^{t+1} = \bm{w}^{(t)} + \hat{\bm{w}}^{(t)}.$
\ENDFOR
\ENSURE Edge weighting $\bm{w}_{\mathrm{out}} = \nicefrac{1}{T} \cdot \sum_{t=0}^{T-1} \bm{w}^{(t)}. $
\end{algorithmic}
\end{algorithm}

\section{Analysis of SDP-based reweighting}\label{sec:sdp}

In this section, we describe and analyze the MMWU algorithm for solving the reweighting SDP problem~\ref{eq:sdp-saddle-point} in Section~\ref{sec:reweighting}. We conclude by proving Lemma~\ref{lemma:reweighting}. 

\subsection{MMWU algorithm for \ref{eq:sdp-saddle-point}}
We present the pseudocode for
the MMWU Algorithm in Algorithm~\ref{alg:MMWU}. 
Our algorithm instantiates the MMWU framework of Arora and Kale~\cite{aroraCombinatorialPrimalDualApproach2016}, where we avoid maintaining full $|\mathcal{V}| \times |\mathcal{V}|$ matrices by relying on the Johnson-Lindenstrauss Lemma~\cite{achlioptasDatabasefriendlyRandomProjections2003} (see lines \ref{line:jl-matrix} and \ref{line:jl-comp}).

At every iteration, the MMWU algorithm produces a candidate solution  $\bm{X}^{(t)} \in \Delta$  to which we respond with a loss matrix $\bm{L}^{(t)} \defeq \bm{L}_{\bm{w}^{(t)}} = \sum_{ij \in \cE_{\mathrm{SR}}} w_{ij}^{(t)} \bm{L}_{ij}$
with $\bm{w}^{(t)} \in \cF$, where $\bm{L}_{ij}\coloneqq (\bm{e}_i - \bm{e}_j)(\bm{e}_i - \bm{e}_j)^\top$.
In this way, the loss $\langle \bm{L}^{(t)}, \bm{X}^{(t)}\rangle $ incurred by the MMWU algorithm at iteration $t$ equals the value of \ref{eq:sdp-saddle-point} on the pair of solutions $(\bm{X}^{(t)}, \bm{w}^{(t)}).$ 
At every iteration $t,$ given $\bm{X}^{(t)}$, our goal is then to choose $\bm{w}^{(t)} \in \cF$ to maximize the loss $\langle \bm{L}^{(t)}, \bm{X}^{(t)} \rangle$. The regret minimization property of MMWU then allows us to turn this per-iteration guarantee into a global guarantee on $\lambda_{n-1}(\sum_{t=0}^{T-1} \bm{L}^{(t)})$.

To maximize the loss of the MMWU algorithm, we choose $\bm{w}^{(t)} \in \cF$ to approximate the best response 
$
\max_{\bm{w} \in \cF} \langle \bm{L}_{\bm{w}}, \bm{X}^{(t)} \rangle. 
$
We describe two algorithms (oracles in the language of \cite{aroraCombinatorialPrimalDualApproach2016}) for approximately solving this task. The first one is based on directly applying a nearly-linear-time packing LP solver. It is described in Appendix~\ref{app:sdp-proofs} and yields the following theorem.
\begin{theorem}\label{thm:LP-oracle}
Given $\bm{X} \in \Delta,$ one can $(1-\epsilon)$-approximate $\max\limits_{\bm{w} \in \cF} \langle \bm{L}_{\bm{w}}, \bm{X} \rangle$ multiplicatively in time $\tilde{O}(\nicefrac{|\cE_{\mathrm{SR}}|}{\epsilon}).$
\end{theorem}

Our second oracle exploits the fact that $\max_{\bm{w} \in \cF} \langle \bm{L}_{\bm{w}}, \bm{X} \rangle = \max_{\bm{w} \in \cF} \langle \bm{c}, \bm{w} \rangle$ is an LP relaxation of the maximum weight $b$-matching problem over $\cG_{\mathrm{SR}}$ with edge weights $\bm{c}$. We can approximate the maximum weight $b$-matching by a greedy procedure: iterate through the edges of $\mathcal{E}_{\mathrm{SR}}$ once in decreasing order of edge gains $c_{ij} = \langle \bm{L}_{ij}, \bm{X} \rangle$, and add an edge so long as both adjacent vertices possess available demand. Though this only achieves a $\nicefrac{1}{2}$-approximation~\cite{mestreGreedyApproximationAlgorithms2006}, the simplicity of the algorithm makes it extremely suitable for implementation. We prove the following theorem in Appendix~\ref{app:sdp-proofs}.
\begin{theorem}\label{thm:greedy-oracle}
Given $\bm{X} \in \Delta,$ one can $\nicefrac{1}{2}$-approximate $\max_{\bm{w} \in \cF} \langle \bm{L}_{\bm{w}}, \bm{X} \rangle$ multiplicatively in time $\tilde{O}(|\mathcal{E}_{\mathrm{SR}}|).$   
\end{theorem}

\subsection{Analysis of Algorithm~\ref{alg:MMWU}}

We are now ready to prove Lemma~\ref{lemma:reweighting}. To analyze the correctness and running time of Algorithm~\ref{alg:MMWU}, we first recall the regret bound achieved by applying MMWU to the vector space $\R^{V}\perp\bm{1}$.
%
\begin{theorem}[Theorem 3.1 \cite{aroraCombinatorialPrimalDualApproach2016}]\label{thm:regret-bound}
Consider a sequence of loss matrices
$\{\bm{L}^{(t)} \in \mathbb{S}^V\}_{t \in [T]}$ 
with $\bm{L}^{(t)} \bm{1} = 0$ for all $t$. 
Let
$$
\bm{W}^{(t)} = \exp\left\{-\eta \sum_{s=0}^{t-1} \bm{L}^{(s)}\right\}, \quad \text{ and } \quad \bm{X}^{(t)} = \frac{\bm{W}^{(t)}}{\langle \Pi_{\perp \bm{1}}, \bm{W}^{(t)} \rangle }.
$$
Then, we have the regret bound:
\begin{equation}\label{eq:regret-bound}
\min_{\bm{X} \in \Delta} \left\langle \sum_{t=0}^{T-1} \bm{L}^{(t)}, \bm{X} \right\rangle \geq 
\sum_{t=0}^{T-1} \langle \bm{L}^{(t)}, \bm{X}^{(t)} \rangle - \eta \sum_{s=0}^{T-1} \langle (\bm{L}^{(t)})^2, \bm{X}^{(t)} \rangle - \frac{\log(n)}{\eta}.   
\end{equation}
\end{theorem}

Next, we use Lemma~\ref{lem:jl} to show that $\bm{V}^{(t)}$, produced by Algorithm~\ref{alg:MMWU}, approximate the MMWU updates $\bm{X}^{(t)}$ in Theorem~\ref{thm:regret-bound} when computing the squared distances $\langle \bm{L}_{ij},\bm{X}^{(t)}\rangle$. This is proven in Appendix~\ref{sec:sdp.apdx}.
\begin{lemma}\label{lem:jl}
 Let $\tilde{\bm{X}}^{(t)}= \bm{V}^{(t)}(\bm{V}^{(t)})^T$ as defined in Algorithm~\ref{alg:MMWU}. Then, $\tilde{\bm{X}}^{(t)} \in \Delta.$ Moreover, for large enough $k \in O(\nicefrac{\log(n)}{\epsilon^2})$, for all pairs $(i,j) \in \mathcal{V} \times \mathcal{V}$, the squared distance $\langle \bm{L}_{ij},\tilde{\bm{X}}^{(t)} \rangle $ is an $\epsilon$-multiplicative approximation to the squared distance $\langle \bm{L}_{ij},\bm{X}^{(t)} \rangle$ with high probability.    
\end{lemma}

We can use the above to establish the algorithm's correctness; we prove the following lemma in Section~\ref{sec:mmwu-correctness-proof}.
\begin{lemma}\label{lem:mmwu-correctness}
With high probability, Algorithm~\ref{alg:MMWU} outputs a reweighting $\bm{w}_\mathrm{out} \in \cF$ of $\cG_{\mathrm{SR}}$ such that
 $$
\min_{\bm{X} \in \Delta}  \langle \bm{L}_{\bm{w}_\mathrm{out}} , \bm{X}\rangle \geq
\begin{cases}
(1 - O(\epsilon)) \cdot \mathrm{OPT}, &\textrm{ if Theorem~\ref{thm:LP-oracle} is used in line~\ref{line:oracle}},\\
(\nicefrac{1}{2} - O(\epsilon)) \cdot \mathrm{OPT}, & \textrm{ if Theorem~\ref{thm:greedy-oracle} is used in line~\ref{line:oracle}}.\\
\end{cases}
 $$
\end{lemma}

Finally, we can bound the running time of Algorithm~\ref{alg:MMWU} using known algorithms for computing the action of the matrix exponential~\cite{aroraCombinatorialPrimalDualApproach2016,orecchiaApproximatingExponentialLanczos2012}. The proof of the next lemma is also given in Appendix~\ref{sec:sdp.apdx}.
\begin{lemma}\label{lem:mmwu-rt} For a constant $\epsilon > 0,$
Algorithm~\ref{alg:MMWU} runs in nearly-linear time in the size of $\cG_{\mathrm{SR}}$.
\end{lemma}

Combining Lemmata~\ref{lem:mmwu-correctness} and~\ref{lem:mmwu-rt} yields our main result on the reweighting of $\cG_{\mathrm{SR}}$, i.e., Lemma~\ref{lemma:reweighting}.
\begin{proof}[Proof of Lemma~\ref{lemma:reweighting}]
 We claim that the output $\bm{w_{\mathrm{out}}}$ of Algorithm~\ref{alg:MMWU} satisfies the required properties.
 The conditions of (\ref{eq:rw-lemma-degree}) and~(\ref{eq:rw-lemma-max}) are immediately satisfied by the fact that $\bm{w_{\mathrm{out}}} \in \cF$, which is proved in Lemma~\ref{lem:mmwu-correctness}.
 The spectral condition in (\ref{eq:rw-lemma-spectral-gap}) follows from the approximation guarantee of Lemma~\ref{lem:mmwu-correctness} and the lower bound~\eqref{eq:OPT}:
 $$
 \lambda_{n-1}\left(\bm{L}_{ \bm{w}_{\mathrm{out}}}\right) = \cdot \min_{\bm{X} \in \Delta}  \langle \bm{L}_{\bm{w}_\textrm{out}} , \bm{X}\rangle  \geq \Omega\left(\mathrm{OPT}\right) \geq \Omega(np).
 $$
 The nearly-linear running time is proved in Lemma~\ref{lem:mmwu-rt}.
\end{proof}

\subsection{Proof of Lemma~\ref{lem:mmwu-correctness}}
\label{sec:mmwu-correctness-proof}

Notice that $\bm{w}_\textrm{out}\in \cF$ as it is the average of elements of $\cF$ and $\cF$ is convex. Moreover, we have $\bm{L}_{\bm{w}_\textrm{out}} = \nicefrac{1}{T} \cdot \sum_{t=0}^{T-1} \bm{L}_{\bm{w}^{(t)}} = \nicefrac{1}{T} \cdot \sum_{t=0}^{T-1} \bm{L}^{(t)}.$
By the regret bound in Equation~\eqref{eq:regret-bound}, we then obtain:
\begin{equation}
\label{eq:main-opt}
\min_{\bm{X} \in \Delta}  \langle \bm{L}_{\bm{w}_\textrm{out}} , \bm{X}\rangle  \geq 
\frac{1}{T} \left(\sum_{t=0}^{T-1} \langle \bm{L}_{\bm{w}^{(t)}}, \bm{X}^{(t)} \rangle\right) - 
\frac{\eta}{T} \left( \sum_{s=0}^{T-1} \langle (\bm{L}_{\bm{w}^{(t)}})^2, \bm{X}^{(t)} \rangle\right) - 
\frac{\log(n)}{\eta T}.    
\end{equation}
We can exploit the positive semi-definiteness of $\bm{L}_{\bm{w}^{(t)}}$ to bound the second term as a function of the first. By the definition of $\cF$, the reweighting of $\cG$ by $\bm{w}^{(t)} \in \cF$ has maximum degree at most $2pn.$ Hence:
$$
\bm{L}_{\bm{w}^{(t)}} \preceq 2\cdot 2pn \cdot  \Pi_{\perp \bm{1}} \textrm{ and } \langle (\bm{L}_{\bm{w}^{(t)}})^2, \bm{X}^{(t)} \rangle \leq 4pn \cdot  \langle \bm{L}_{\bm{w}^{(t)}}, \bm{X}^{(t)} \rangle.
$$
We can now rewrite Equation~\eqref{eq:main-opt} as:
$$
\min_{\bm{X} \in \Delta}  \langle \bm{L}_{\bm{w}_\textrm{out}} , \bm{X}\rangle  \geq 
\left(\frac{1}{T} - \frac{4pn \cdot \eta}{T}\right) \left(\sum_{t=0}^{T-1} \langle \bm{L}_{\bm{w}^{(t)}}, \bm{X}^{(t)} \rangle\right)- 
\frac{\log(n)}{\eta T}.  
$$
For all iterations  $t$, we have:
\begin{align*}
\langle \bm{L}_{\bm{w}^{(t)}}, \bm{X}^{(t)} \rangle \geq
(1 - \epsilon)\langle \bm{L}_{\bm{w}^{(t)}}, \tilde{\bm{X}}^{(t)} \rangle =
(1- \epsilon) \sum_{ij \in \cE_{\mathrm{SR}}} w_{ij}^{(t)} \langle \bm{L}_{ij} ,\bm{V}^{(t)}(\bm{V}^{(t)})^T\rangle 
 =
(1-\epsilon) \max_{\bm{w} \in \cF} \langle \bm{L}_{\bm{w}}, \tilde{\bm{X}}^{(t)} \rangle, 
\end{align*}
where the first inequality follows from Lemma~\ref{lem:jl} and the last equality follows from Lines~\ref{line:gains} and~\ref{line:oracle} in Algorithm~\ref{alg:MMWU}.
As $\tilde{\bm{X}}^{(t)} \in \Delta$, the maximum in the last expression has value at least $\mathrm{OPT}$. Hence, for the two oracles of Theorems~\ref{thm:LP-oracle} and~\ref{thm:greedy-oracle}, we have:
$$
\mathrm{Theorem~\ref{thm:LP-oracle}:} \langle \bm{L}_{\bm{w}^{(t)}}, \bm{X}^{(t)} \rangle \geq (1 - \epsilon)^2 \cdot \mathrm{OPT} \;
\mathrm{ and~Theorem~\ref{thm:greedy-oracle}:}
\langle \bm{L}_{\bm{w}^{(t)}}, \bm{X}^{(t)} \rangle \geq \frac{(1 - \epsilon)}{2} \cdot \mathrm{OPT}.
$$
Therefore:
$$
\min_{\bm{X} \in \Delta}  \langle \bm{L}_{\bm{w}_\textrm{out}} , \bm{X}\rangle  \geq \begin{cases}
(1 - 4pn \cdot \eta) \cdot (1-\epsilon)^2 \cdot \mathrm{OPT} - \frac{\log(n)}{\eta T}\\
(1 - 4pn\cdot \eta) \cdot \frac{(1-\epsilon)}{2} \cdot \mathrm{OPT} - \frac{\log(n)}{\eta T}\\
\end{cases}
$$
Substituting the definitions of $\eta=\nicefrac{\epsilon}{4pn}$  yields:
$$
\min_{\bm{X} \in \Delta}  \langle \bm{L}_{\bm{w}_\textrm{out}} , \bm{X}\rangle  \geq \begin{cases}
(1-\epsilon)^3 \cdot \mathrm{OPT} - \frac{4pn \log(n)}{\epsilon T}\\
 \frac{(1-\epsilon)^2}{2} \cdot \mathrm{OPT} - \frac{4 pn \log(n)}{\epsilon T}\\
\end{cases}
$$
By the lower bound~\eqref{eq:OPT}, the last term in both expressions can be upper bound by $\nicefrac{8\mathrm{OPT} \log(n)}{\epsilon T}.$ The statement of the lemma follows from the definition $T= \nicefrac{8\log(n)}{\epsilon^2}.$







\section{Discussion}
In this paper we consider top-$K$ ranking with a monotone adversary. We carefully construct a weighted MLE that achieves the optimal $\ell_{\infty}$ estimation error and
top-$K$ recovery sample complexity. This leaves open quite a few interesting directions. We single out several of them below. 
\begin{itemize}
\item \emph{Shaving the $\log^2(n)$ factor.} Compared to the results for uniform sampling, ours requires an extra assumption that $npL\gtrsim\log^3(n)$. While this is optimal for a wide range of $\Delta_{K}$, it does incur an extra $\log^2(n)$ factor when $1/\log(n) \lesssim \Delta_{K} \lesssim 1$. We do not expect this to be the fundamental gap between uniform sampling and semi-random sampling. Successfully shaving this extra log factor will potentially bring more insights to understanding the geometry of the comparison graph on the ranking problem.

\item \emph{Is weighted MLE necessary?} While our approach relies on the weighted MLE, it is not clear whether the unweighted vanilla MLE succeeds or not with a monotone adversary. In fact, in Appendix~\ref{sec:cluster}, we present a specific example of semi-random sampling where the spectral properties of the semi-random graph are drastically different from those of random graph, yet the vanilla MLE still succeeds.


\item \emph{Extension to general graphs}. Theorem 1 in \cite{chen2023ranking}
 and Theorem~\ref{thm:weighted_MLE} in our paper both consider general
comparison graphs. Our analysis can recover a good rate with small
sample complexity when the spectrum of the comparison graph has small
dynamic range, i.e. $\lambda_{1}/\lambda_{n-1}$ is small. Our reweighting procedure indicates that any comparison graph which can be reweighted to match this spectral condition will yield small sample complexity. It is an interesting question to explore the connection between the reweighting SDP and similar SDPs used in the approximation of general partitioning objectives~\cite{lauCheegerInequalitiesDirected2023} to find novel combinatorial or geometric properties enabling the success of the weighted MLE.

\item \emph{Extensions to other ranking models.} In this paper we focus on
one of the most popular model in ranking, namely the BTL model. It
is certainly interesting to see whether our algorithm and
analysis for semi-random sampling can be extended to other models,
for instance the Thurstone model \cite{thurstone1927law}, the Plackett-Luce model \cite{luce2005individual}, and other models for multi-way comparisons~\cite{fan2022ranking,fan2023spectral}.
\end{itemize}

\subsection*{Acknowledgements}
AC is supported by NSF DGE 2140001. LO is supported by NSF CAREER 1943510. CM was partially supported by the National Science Foundation via grant DMS-2311127. 

\bibliographystyle{alpha}
\bibliography{All-of-Bibs,lorenzo}

\appendix

\section{Auxiliary lemmas}
This section collects several auxiliary lemmas we use in the proofs of our main results. 
\begin{lemma}[Range of $z_{ij}$]\label{lemma:z_range} Recall that
\[
z_{ij}=\frac{e^{\theta_{i}^{\star}}e^{\theta_{j}^{\star}}}{(e^{\theta_{i}^{\star}}+e^{\theta_{j}^{\star}})^{2}}=\frac{e^{\theta_{i}^{\star}-\theta_{j}^{\star}}}{(1+e^{\theta_{i}^{\star}-\theta_{j}^{\star}})^{2}}.
\]
We have for any $i,j$,  
\[
\frac{1}{4\kappa}\le z_{ij}\le\frac{1}{4}.
\]

\end{lemma}\begin{proof}

The function $f(x)=x/(1+x)^{2}$ has derivative $(1-x^{2})/(1+x)^{4}$ so it is increasing in $(0, 1)$ and decreasing in $(1,\infty)$.
Furthermore from the definition of $\kappa$, $|\theta_{i}^{\star}-\theta_{j}^{\star}|\le\log(\kappa)$.
Then we have 
\[
\frac{1}{4\kappa}\le \min\left\{f(\kappa), f(1/\kappa) \right\}\le z_{ij}\le f(1)=\frac{1}{4}.
\]
This completes the proof. 
\end{proof}

\begin{lemma}[spectral gap of a Erd\H{o}s-R\'{e}nyi graph]\label{lemma:spectral_ER} Let $\mathcal{G} = ([n], \mathcal{E})\sim G(n, p)$ be an Erd\H{o}s-R\'{e}nyi graph with $n$ vertices and edge probability $p$. Let $\bm{L}$ be its corresponding graph Laplacian matrix. Let $d_{\max}$ be the maximum degree of the vertices. Suppose that $np\ge C\log (n)$ for some large enough constant $C>0$, then with probability at least $1-n^{-10}$,
\begin{align*}
  \lambda_{n-1}(\bm{L}) &\ge \frac{np}{2}  \\
  d_{\max} &\le 2np.
\end{align*}

\end{lemma}
\begin{proof}
    See, for instance, Section~5.3.3 of \cite{tropp2015matrix}.
\end{proof}

\begin{lemma}[\cite{spielman2007spectral}] \label{lemma:conductance_eigen}
Let $\mathcal{G}$ be a weighted graph and $\Phi(\mathcal{G})$ be its conductance (see Definition~\ref{def:conductance}). Let  $\bm{L}$ be its graph Laplacian matrix. Let $\bm{D}$ be the diagonal matrix with the weighted vertex degrees as the entries. Then

\[
\Phi(\mathcal{G})\ge\frac{1}{2}\lambda_{n-1}(\bm{D}^{-1/2}\bm{L}\bm{D}^{-1/2}).
\]

\end{lemma}



\begin{lemma}[Rayleigh's monotonicity law, \cite{bollobas1998modern} Corollary 7 in Ch.IX]\label{lemma:monotonicity}
Let $\mathcal{G}=(\mathcal{V},\mathcal{E})$ be a graph with weights
$\{w_{ij}\}$ on $\mathcal{E}$, and $\widetilde{\mathcal{G}}=(\mathcal{V},\widetilde{\mathcal{E}})$
be a graph with weights $\{\widetilde{w}_{ij}\}$ on $\widetilde{\mathcal{E}}$.
Suppose that $\mathcal{E}\subset\widetilde{\mathcal{E}}$ and $w_{ij}\le\widetilde{w}_{ij}$
for any $(i,j)\in\mathcal{E}$, then for any $(k,l)\in \mathcal{V}^2$, the effective resistance satisfies
\[
\bm{\Omega}_{kl}(\bm{L}_{\mathcal{G}_{w}})\le\bm{\Omega}_{kl}(\bm{L}_{\widetilde{\mathcal{G}}_{\widetilde{w}}}).
\]

\end{lemma}

\begin{lemma}[A quantitative version of Sylvester's law of inertia, \cite{ostrowski1959quantitative}]
\label{lemma:sylvester} For any real symmetric matrix $\bm{A}\in\mathbb{R}^{n\times n}$
and $\bm{S}\in\mathbb{R}^{n\times n}$ be a non-singular matrix. Then
for any $i\in[n]$, $\lambda_{i}(\bm{S}\bm{A}\bm{S}^{\top})$ lies
between $\lambda_{i}(\bm{A})\lambda_{1}(\bm{S}^{\top}\bm{S})$ and
$\lambda_{i}(\bm{A})\lambda_{n}(\bm{S}^{\top}\bm{S})$.

\end{lemma}

\begin{lemma}
 \label{lemma:d_min}
 Let $\mathcal{G} = (\mathcal{V}, \mathcal{E}, \{w_{ij}\}_{(i,j)\in\mathcal{E}} )$ be a weighted graph. Recall $d_{\min} = \min_{i\in[n]}\sum_{j:(i,j)\in\mathcal{E}} w_{ij}$ is the minimum weighted degree and $\bm{L}_w = \sum_{i>j}w_{ij} (\bm{e}_i - \bm{e}_j)(\bm{e}_i - \bm{e}_j)^\top$ is the weighted graph Laplacian. Then 
    \[d_{\min}\ge \frac{1}{2}\lambda_{n-1}(\bm{L}_w).\]
\end{lemma}
\begin{proof}
    Let $k\in [n]$ be an arbitrary vertex. Let $\bm{v}\in \mathbb{R}^n$ be a vertex defined by $v_k = 1$ and $v_j = -1/(n-1)$ for all $j\neq i$. It is easy to see that $\bm{v}^\top \bm{1}_n = 0$. Moreover, for any $i,j\neq k$, $(\bm{e}_{k} - \bm{e_j})^\top \bm{v} = n/(n-1)$ and $(\bm{e}_{i} - \bm{e_j})^\top \bm{v} = 0$. By the definition of $\lambda_{n-1}(\bm{L}_w)$, we have that 
\begin{align*}
        \lambda_{n-1}(\bm{L}_w) &= \min_{\bm{u}\in\mathbb{R}^n:\bm{u}^\top  \bm{1}_n = 0} \frac{\bm{u}^\top \bm{L}_w \bm{u}}{\|\bm{u}\|^2}\\
        &\le \frac{\bm{v}^\top \bm{L}_w \bm{v}}{\|\bm{v}\|^2}\\
        &=\frac{\sum_{j:j\neq k} w_{kj} \left(\frac{n}{n-1}\right)^2}{n/(n-1)} \\
        &\le 2\sum_{j:j\neq k} w_{kj}
\end{align*} 
as long as $n\ge 2$. This holds for all $k\in [n]$ so 
$ d_{\min}\ge \frac{1}{2}\lambda_{n-1}(\bm{L}_w)$.
\end{proof}

\section{Proof of Lemma~\ref{thm:weighted_MLE}\label{sec:proof-weighted-MLE-B-Q}}
We start by defining some useful notations. Let $\epsilon_{ij}^{(l)}=y_{ji}^{(l)}-\sigma(\theta_{i}^{\star}-\theta_{j}^{\star})$
where $\sigma(x)=e^{x}/(1+e^{x})$ is the sigmoid function. Let $\bm{B}\coloneqq\left[\cdots,\sqrt{w_{ij}z_{ij}}(\bm{e}_{i}-\bm{e}_{j}),\cdots\right]\in\mathbb{R}^{n\times n(n-1)L/2}$
(each entry repeats for $L$ times), $\widehat{\bm{\epsilon}}=\left[\cdots,\sqrt{w_{ij}/z_{ij}}\epsilon_{ji}^{(l)},\cdots\right]\in\mathbb{R}^{n(n-1)L/2}$
and $\bm{\delta}^{t}\coloneqq\bm{\theta}^{t}-\bm{\theta}^{\star}$.
Observe that 
\begin{equation}
\bm{B}\bm{B}^{\top}=L\sum_{(i,j):i>j}w_{ij}z_{ij}(\bm{e}_{i}-\bm{e}_{j})(\bm{e}_{i}-\bm{e}_{j})^{\top}=L\cdot\bm{L}_{wz}.\label{eq:BB_top}
\end{equation}

Inspired by~\cite{chen2023ranking}, we analyze the weighted MLE by studying the preconditioned
gradient descent starting from the ground truth. Let $\bm{\theta}_{0}=\bm{\theta}^{\star}$
be the starting point and $\eta>0$ be the step size. 
\begin{equation}
\bm{\theta}^{t+1}=\bm{\theta}^{t}-\eta\bm{L}_{wz}^{\dagger}\nabla\mathcal{L}_{w}(\bm{\theta}^{t}).\label{eq:precondGD}
\end{equation}
It is worth noting that because the preconditioned gradient descent starts at ground
truth and the gradient is of form
\[
\nabla\mathcal{L}_{w}(\bm{\theta})=\frac{1}{L}\sum_{i,j:i>j}\sum_{l=1}^{L}w_{ij}\left\{ -y_{ji}^{(l)}+\frac{e^{\theta_{i}}}{e^{\theta_{i}}+e^{\theta_{j}}}\right\} (\bm{e}_{i}-\bm{e}_{j}),
\]
we have $\bm{1}_{n}^{\top}\bm{\theta}^{t}=0$ for all $t\ge0$. We
break the proof of Lemma~\ref{thm:weighted_MLE} into the following
lemmas. The first states that $\bm{\theta}^{t}$ stays close to the
ground truth in $\ell_{\infty}$ distance, and the second states that
it converges to the weighted MLE.

\begin{lemma}\label{lemma:weighted_MLE_iter}Instate the assumptions
of Lemma~\ref{thm:weighted_MLE}. With probability at least $1-n^{-10}$,
(\ref{eq:subGaussian_concen}) is satisfied and for any $k,l$ and
iteration $t\ge0$,
\begin{equation}
\left|\left(\theta_{k}^{t}-\theta_{l}^{t}\right)-\left(\theta_{k}^{\star}-\theta_{l}^{\star}\right)\right|\le B_{kl}.\label{eq:B_itr_t}
\end{equation}
\end{lemma}
For any $\bm{x}\in\mathbb{R}$ and $\bm{Z}\in\mathbb{R}^{n\times n}$, let
$\|\bm{x}\|_{\bm{Z}}\coloneqq(\bm{x}^{\top}\bm{Z}\bm{x})^{1/2}.$
\begin{lemma}\label{lemma:preconGD_converge}Instate the assumptions
of Lemma~\ref{thm:weighted_MLE}. On the event that (\ref{eq:subGaussian_concen})
is satisfied, the following holds:
\begin{enumerate}
\item There exists a unique minimizer $\widehat{\bm{\theta}}$ of (\ref{eq:MLE_loss}).
\item There exists some $\alpha_{1},\alpha_{2}\in\mathbb{R}$ such that
$0<\alpha_{1}\le\alpha_{2}$ and for any $t\in\mathbb{N}$, 
\end{enumerate}
\[
\|\bm{\theta}^{t}-\widehat{\bm{\theta}}\|_{\bm{L}_{wz}}\le(1-\eta\alpha_{1})^{t}\|\bm{\theta}^{0}-\widehat{\bm{\theta}}\|_{\bm{L}_{wz}},
\]
provided that $0<\eta\le1/\alpha_{2}$.

\end{lemma}

We will prove these two lemmas in Section~\ref{subsec:Proof_stay_close}
and Section \ref{subsec:Proof_convergence}. Their proofs are similar
to the proof of Theorem~1 in \cite{chen2023ranking}. As we are studying
weighted MLE instead of the unweighted MLE used in \cite{chen2023ranking},
we redo the proofs and add necessary modifications. 

\paragraph{Proof of Lemma~\ref{thm:weighted_MLE}.}

It is easy to see that Lemma~\ref{lemma:preconGD_converge} shows
the preconditioned gradient descent iterates $\bm{\theta}^{t}$ converge
to a unique solution $\widehat{\bm{\theta}}$. Combining this with
Lemma~\ref{lemma:weighted_MLE_iter}, we have
\[
\left|\left(\widehat{\theta}_{k}-\widehat{\theta}_{l}\right)-\left(\theta_{k}^{\star}-\theta_{l}^{\star}\right)\right|\le B_{kl}.
\]
As $\bm{1}^\top_n \bm{\theta}^\star  = \bm{1}^\top_n \widehat{\bm{\theta}} = 0$, for any $i\in[n]$,
\begin{align*}
\left|\widehat{\theta}_{i}-\theta_{i}^{\star}\right| & =\left|\frac{1}{n}\sum_{j=1}^{n}\left(\widehat{\theta}_{i}-\widehat{\theta}_{j}\right)-\left(\theta_{i}^{\star}-\theta_{j}^{\star}\right)\right|\\
 & \le\frac{1}{n}\sum_{j=1}^{n}\left|\left(\widehat{\theta}_{i}-\widehat{\theta}_{j}\right)-\left(\theta_{i}^{\star}-\theta_{j}^{\star}\right)\right|\\
 & \le \max_{k,l}B_{kl}.
\end{align*}
It remains to show the exact recovery of the top-$K$ items. Recall
$\theta_{1}^{\star}\ge\cdots\ge\theta_{K}^{\star}>\theta_{K+1}^{\star}\ge\cdots\ge\theta_{n}^{\star}$
by assumption. It suffices to show $\widehat{\theta}_{i}-\widehat{\theta}_{j}>0$
for any $i\le K$, $j>K$. Let $i\le K,j>K$,
\begin{align*}
\widehat{\theta}_{i}-\widehat{\theta}_{j} & \ge\left(\theta_{i}^{\star}-\theta_{j}^{\star}\right)-\left|\left(\widehat{\theta}_{i}-\widehat{\theta}_{j}\right)-\left(\theta_{i}^{\star}-\theta_{j}^{\star}\right)\right|\ge\Delta_{K}-B_{ij}.
\end{align*}
Then $\widehat{\theta}_{i}-\widehat{\theta}_{j}>0$ as long as 
\[
 \max_{k,l}B_{kl} \le \Delta_K / 2.
\]
Now the proof of Lemma~\ref{thm:weighted_MLE} is completed.

\subsection{Proof of Lemma~\ref{lemma:weighted_MLE_iter}\label{subsec:Proof_stay_close}}

We will prove this lemma by induction on $t$. Since preconditioned
gradient descent starts at ground truth, the base case of $t=0$ is
trivial. We will now study the dynamics of (\ref{eq:precondGD})
to prove the induction step. Using the definition (\ref{eq:MLE_loss})
we can compute the gradient and Hessian of the loss function: 
\[
\nabla\mathcal{L}(\bm{\theta})=\frac{1}{L}\sum_{i,j:i>j}\sum_{l=1}^{L}w_{ij}\left\{ -y_{ji}^{(l)}+\frac{e^{\theta_{i}}}{e^{\theta_{i}}+e^{\theta_{j}}}\right\} (\bm{e}_{i}-\bm{e}_{j});
\]
\[
\nabla^{2}\mathcal{L}(\bm{\theta})=\sum_{i,j:i>j}w_{ij}\frac{e^{\theta_{i}}e^{\theta_{j}}}{\left(e^{\theta_{i}}+e^{\theta_{j}}\right)^{2}}(\bm{e}_{i}-\bm{e}_{j})(\bm{e}_{i}-\bm{e}_{j})^{\top}.
\]
Applying Taylor's expansion on $\nabla\mathcal{L}(\bm{\theta}^{t})$,
we have
\begin{align*}
\nabla\mathcal{L}(\bm{\theta}^{t}) & =\frac{1}{L}\sum_{i,j:i>j}\sum_{l=1}^{L}w_{ij}\left(\left(-y_{ji}^{(l)}+\sigma(\theta_{i}^{t}-\theta_{j}^{t})\right)(\bm{e}_{i}-\bm{e}_{j})\right)\\
 & =\frac{1}{L}\sum_{i,j:i>j}\sum_{l=1}^{L}w_{ij}\left(\left(-\epsilon_{ji}^{(l)}-\sigma(\theta_{i}^{\star}-\theta_{j}^{\star})+\sigma(\theta_{i}^{t}-\theta_{j}^{t})\right)(\bm{e}_{i}-\bm{e}_{j})\right)\\
 & =\frac{1}{L}\sum_{i,j:i>j}\sum_{l=1}^{L}w_{ij}\left(\left(-\epsilon_{ji}^{(l)}+\sigma'(\theta_{i}^{\star}-\theta_{j}^{\star})(\delta_{i}^{t}-\delta_{j}^{t})+\frac{1}{2}\sigma''(\zeta_{ij}^{t})(\delta_{i}^{t}-\delta_{j}^{t})^{2}\right)(\bm{e}_{i}-\bm{e}_{j})\right).
\end{align*}
Here for all $i,j\in[n]$, $\delta_{i}^{t}\coloneqq\theta_{i}^{t}-\theta_{i}^{\star}$
and $\zeta_{ij}^{t}\in\mathbb{R}$ is some number that lies between
$\theta_{i}^{\star}-\theta_{j}^{\star}$ and $\theta_{i}^{t}-\theta_{j}^{t}$.
As $\sigma'(\theta_{i}^{\star}-\theta_{j}^{\star})=z_{ij}$ and $\delta_{i}^{t}-\delta_{j}^{t}=(\bm{e}_{i}-\bm{e}_{j})^{\top}\bm{\delta}$,
we can rewrite the above formula as
\[
\nabla\mathcal{L}(\bm{\theta}^{t})=\frac{1}{L}\left(L\cdot\bm{L}_{wz}\bm{\delta}^{t}-\bm{B}\widehat{\bm{\epsilon}}+L\cdot\bm{r}^{t}\right),
\]
where $\bm{r}^{t}\coloneqq\sum_{i,j:i>j}w_{ij}[\frac{1}{2}\sigma''(\zeta_{ij}^{t})(\delta_{i}^{t}-\delta_{j}^{t})^{2}(\bm{e}_{i}-\bm{e}_{j})]$.
Putting this in (\ref{eq:precondGD}), we have
\begin{equation}
\bm{\delta}^{t+1}=\left(1-\eta\right)\bm{\delta}^{t}-\frac{\eta}{L}\left(\bm{L}_{wz}^{\dagger}\bm{B}\widehat{\bm{\epsilon}}-L\cdot\bm{L}_{wz}^{\dagger}\bm{r}^{t}\right).\label{eq:precGD_itr_t}
\end{equation}
Now consider $\delta_{k}^{t}-\delta_{l}^{t}=(\bm{e}_{k}-\bm{e}_{l})^{\top}\bm{\delta}^{t}$,
we have
\begin{equation}
\delta_{k}^{t+1}-\delta_{l}^{t+1}=\left(1-\eta\right)\left(\delta_{k}^{t}-\delta_{l}^{t}\right)-\frac{\eta}{L}(\bm{e}_{k}-\bm{e}_{l})^{\top}\left(\bm{L}_{wz}^{\dagger}\bm{B}\widehat{\bm{\epsilon}}-L\cdot\bm{L}_{wz}^{\dagger}\bm{r}^{t}\right).\label{eq:precGD_kl}
\end{equation}
We now control the size of $(\bm{e}_{k}-\bm{e}_{l})^{\top}\bm{L}_{wz}^{\dagger}\bm{B}\widehat{\bm{\epsilon}}$
and $(\bm{e}_{k}-\bm{e}_{l})^{\top}\bm{L}_{wz}^{\dagger}\bm{r}^{t}$
separately. 

\paragraph{Controlling $(\bm{e}_{k}-\bm{e}_{l})^{\top}\bm{L}_{wz}^{\dagger}\bm{B}\widehat{\bm{\epsilon}}$.}

We rewrite this term with

\[
(\bm{e}_{k}-\bm{e}_{l})^{\top}\bm{L}_{wz}^{\dagger}\bm{B}\widehat{\bm{\epsilon}}=\langle\bm{B}^{\top}\bm{L}_{wz}^{\dagger}(\bm{e}_{k}-\bm{e}_{l}),\widehat{\bm{\epsilon}}\rangle.
\]
By Lemma~\ref{lemma:z_range}, $z_{ij}\ge1/(4\kappa)$ so each entry
of $\widehat{\bm{\epsilon}}$ is of sub-Gaussian norm $\sqrt{w_{ij}/z_{ij}}\le\sqrt{4\kappa\max_{ij}w_{ij}}$.
Then $\langle\bm{B}^{\top}\bm{L}_{wz}^{\dagger}(\bm{e}_{k}-\bm{e}_{l})^{\top},\widehat{\bm{\epsilon}}\rangle$
has sub-Gaussian norm 
\[
\|\bm{B}^{\top}\bm{L}_{wz}^{\dagger}(\bm{e}_{k}-\bm{e}_{l})\|_{2}\sqrt{4\kappa\max_{ij}w_{ij}}\le\sqrt{4\kappa L\max_{ij}w_{ij}\Omega_{kl}(\bm{L}_{wz})} ,
\]
where the inequality comes from 
\begin{align*}
\|\bm{B}^{\top}\bm{L}_{wz}^{\dagger}(\bm{e}_{k}-\bm{e}_{l})^{\top}\|_{2}^{2} & =(\bm{e}_{k}-\bm{e}_{l})\bm{L}_{wz}^{\dagger}\bm{B}\bm{B}^{\top}\bm{L}_{wz}^{\dagger}(\bm{e}_{k}-\bm{e}_{l})\\
 & =L\cdot(\bm{e}_{k}-\bm{e}_{l})\bm{L}_{wz}^{\dagger}(\bm{e}_{k}-\bm{e}_{l})\\
 & =L\cdot\Omega_{kl}(\bm{L}_{wz}).
\end{align*}
The second line follows from (\ref{eq:BB_top}). Taking a union bound
on concentration of sub-Gaussian random variables (see for instance
Section 2.5 in\cite{vershynin2018high}), we have that with probability
at least $1-n^{-10}$, for all $k,l\in[n]$,
\begin{equation}
\left|(\bm{e}_{k}-\bm{e}_{l})^{\top}\bm{L}_{wz}^{\dagger}\bm{B}\hat{\bm{\epsilon}}\right|\le C\sqrt{\kappa L\max_{i,j}w_{ij}\cdot\bm{\Omega}_{kl}(\bm{L}_{wz})\log(n)}\le\frac{1}{2}L\cdot B_{kl}\label{eq:subGaussian_concen}
\end{equation}
 for some constant $C$.

\paragraph{Controlling $(\bm{e}_{k}-\bm{e}_{l})^{\top}\bm{L}_{wz}^{\dagger}\bm{r}^{t}$.}

We expand the term to get
\begin{align*}
\left|(\bm{e}_{k}-\bm{e}_{l})^{\top}\bm{L}_{wz}^{\dagger}\bm{r}^{t}\right| & =\left|\sum_{i,j:i>j}w_{ij}\left[\frac{1}{2}\sigma''(\zeta_{ij}^{t})(\delta_{i}^{t}-\delta_{j}^{t})^{2}(\bm{e}_{k}-\bm{e}_{l})^{\top}\bm{L}_{wz}^{\dagger}(\bm{e}_{i}-\bm{e}_{j})\right]\right|\\
 & \le\frac{1}{8}\sum_{i,j:i>j}w_{ij}(\delta_{i}^{t}-\delta_{j}^{t})^{2}\left|(\bm{e}_{k}-\bm{e}_{l})^{\top}\bm{L}_{wz}^{\dagger}(\bm{e}_{i}-\bm{e}_{j})\right|.
\end{align*}
For the inequality here we use the fact that $\sigma''(\zeta)\le1/4$
for any $\zeta\in\mathbb{R}$.

Now show the induction step. Assume (\ref{eq:B_itr_t}) holds for
iteration $t$, then
\begin{align*}
\left|(\bm{e}_{k}-\bm{e}_{l})^{\top}\bm{L}_{wz}^{\dagger}\bm{r}^{t}\right| & \le\frac{1}{8}\sum_{i,j:i>j}w_{ij}(\delta_{i}^{t}-\delta_{j}^{t})^{2}\left|(\bm{e}_{k}-\bm{e}_{l})^{\top}\bm{L}_{wz}^{\dagger}(\bm{e}_{i}-\bm{e}_{j})\right|\\
 & \le\frac{1}{8}\sum_{i,j:i>j}w_{ij}B_{ij}^{2}\left|(\bm{e}_{k}-\bm{e}_{l})^{\top}\bm{L}_{wz}^{\dagger}(\bm{e}_{i}-\bm{e}_{j})\right|\\
 & =\frac{1}{8}Q_{kl}.
\end{align*}
Substituting this and (\ref{eq:subGaussian_concen}) into (\ref{eq:precGD_kl}),
\begin{align*}
\left|\delta_{k}^{t+1}-\delta_{l}^{t+1}\right| & \le\left(1-\eta\right)\left|\delta_{k}^{t}-\delta_{l}^{t}\right|+\frac{\eta}{L}\left(\left|(\bm{e}_{k}-\bm{e}_{l})^{\top}\bm{L}_{wz}^{\dagger}\bm{r}^{t}\right|+L\cdot\left|(\bm{e}_{k}-\bm{e}_{l})^{\top}\bm{L}_{wz}^{\dagger}\bm{B}\hat{\bm{\epsilon}}\right|\right)\\
 & \le\left(1-\eta\right)B_{kl}+\frac{\eta}{L}\left(\frac{1}{2}L\cdot B_{kl}+\frac{1}{8}L\cdot Q_{kl}\right)\\
 & \le B_{kl}
\end{align*}
as long as $Q_{kl}\le4B_{kl}$.

\subsection{Proof of Lemma~\ref{lemma:preconGD_converge}\label{subsec:Proof_convergence}}

The proof of this follows a similar strategy to Lemma~1 in \cite{chen2023ranking}.
We will only prove the first part here, i.e. the existence and uniqueness
of a solution for (\ref{eq:MLE_loss}), as the proof for the second part
is the same as \cite{chen2023ranking}. We first make the following
claim that we will prove at the end of this subsection. This claim
is analogous to a classical result in \cite{ford1957solution}, which
states the same thing but with unweighted MLE.

\begin{claim}\label{claim:exist_unique_MLE} Instate the assumptions
of Lemma~\ref{lemma:preconGD_converge}. Suppose that for any disjoint
partition $\mathcal{S}_{1}\cup\mathcal{S}_{2}=[n]$, there exists
some $i\in\mathcal{S}_{1},j\in\mathcal{S}_{2}$ such that $w_{ij}>0$ 
and $y_{ji}^{(l)}=1$ for some $1\le l\le L$. Then there exists a minimizer
$\widehat{\bm{\theta}}$ of (\ref{eq:MLE_loss}) that satisfies $\bm{1}_{n}^{\top}\widehat{\bm{\theta}}=0$.
Furthermore it is unique in the sense that no other minimizer $\bm{\theta}$
satisfies $\bm{1}_{n}^{\top}\bm{\theta}=0$.

\end{claim}

Now it suffices to show that the condition of this claim is true. Suppose for the sake of contradiction that it is false. That is there exists some disjoint partition $\mathcal{S}_{1}\cup\mathcal{S}_{2}=[n]$,
such that for all $i\in\mathcal{S}_{1},j\in\mathcal{S}_{2}$ with
$w_{ij}>0$, $y_{ji}=(1/L)\sum_{l=1}^{L}=y_{ji}^{(l)}=0$. By Lemma~\ref{lemma:weighted_MLE_iter} and the fact that $\bm{1}^\top_n \bm{\theta}^t = 0$
for any $t\ge0$, $\bm{\theta}^{t}\in B_{R}\coloneqq\{\bm{\theta}:\|\bm{\theta}\|_{\infty}\le R\}$
for some scalar $R$ that depends on the observations. Now we consider
the minimum size of the gradient in this close ball. Recall
\begin{align*}
\nabla\mathcal{L}(\bm{\theta}) & =\sum_{i,j:i>j}w_{ij}\left\{ -y_{ji}+\frac{e^{\theta_{i}}}{e^{\theta_{i}}+e^{\theta_{j}}}\right\} (\bm{e}_{i}-\bm{e}_{j}).
\end{align*}
Consider a vector $\bm{v}=\bm{1}_{\mathcal{S}_{1}}$ that is $1$
on all entries for $\mathcal{S}_{1}$ and 0 otherwise. It is easy
to see that$(\bm{e}_{i}-\bm{e}_{j})^{\top}\bm{v}=0$ if $i,j\in\mathcal{S}_{1}$
or $i,j\in\mathcal{S}_{2}$; $(\bm{e}_{i}-\bm{e}_{j})^{\top}\bm{v}=1$ if $i\in\mathcal{S}_{1},j\in\mathcal{S}_{2}$. Then for any $\bm{\theta}\in B_R$, one can rearrange the summation to reach
\begin{align*}
\nabla\mathcal{L}(\bm{\theta})^{\top}\bm{v} & =\sum_{i\in\mathcal{S}_{1},j\in\mathcal{S}_{2}}w_{ij}\frac{e^{\theta_{i}}}{e^{\theta_{i}}+e^{\theta_{j}}}(\bm{e}_{i}-\bm{e}_{j})^{\top}\bm{v}\\
 & =\sum_{i\in\mathcal{S}_{1},j\in\mathcal{S}_{2}}w_{ij}\frac{e^{\theta_{i}}}{e^{\theta_{i}}+e^{\theta_{j}}}\\
 & \overset{\text{(i)}}{\ge}\sum_{i\in\mathcal{S}_{1},j\in\mathcal{S}_{2}}w_{ij}\frac{e^{-R}}{e^{-R}+e^{R}}\\
 & \overset{\text{(ii)}}{\ge}\frac{1}{2}e^{-2R}\min\{w_{ij}:w_{ij}>0\}.
\end{align*}
Here (i) holds since $\bm{\theta}\in B_R$ and the function $(x,y)\mapsto e^{x}/(e^{x}+e^{y})$ is increasing in $x$ and decreasing in $y$; (ii) holds since we assume the weighted graph to be
connected by edges with non-zero weights. Taking infimum over $\bm{\theta}$
in $B_{R}$, we have $\inf_{\bm{\theta}\in B_{R}}\|\nabla\mathcal{L}(\bm{\theta})\|>0$. 

It now suffices to show $\|\nabla\mathcal{L}(\bm{\theta}^{t})\|\rightarrow0$
as $t\rightarrow\infty$ which contradicts the claim that $\inf_{\bm{\theta}\in B_{R}}\|\nabla\mathcal{L}(\bm{\theta})\|>0$.
Recall $\bm{L}_{w}$ is the Laplacian matrix of the weighted graph defined as $\sum_{(i,j):i>j}w_{ij}(\bm{e}_{i}-\bm{e}_{j})(\bm{e}_{i}-\bm{e}_{j})^{\top}$.
Since the graph is connected by edges with non-zero weights, $\lambda_{n-1}(\bm{L}_{w})>0$. Consider the Hessian
\[
\nabla^{2}\mathcal{L}(\bm{\theta})=\sum_{(i,j):i>j}w_{ij}\frac{e^{\theta_{i}}e^{\theta_{j}}}{(e^{\theta_{i}}+e^{\theta_{j}})^{2}}(\bm{e}_{i}-\bm{e}_{j})(\bm{e}_{i}-\bm{e}_{j})^{\top}.
\]
For any unit vector $\bm{v}\in\mathbb{R}^{n}$ such that $\bm{v}^{\top}\bm{1}_{n}=0$,
\begin{align*}
\bm{v}^{\top}\nabla^{2}\mathcal{L}(\bm{\theta})\bm{v} & =\sum_{(i,j):i>j}w_{ij}\frac{e^{\theta_{i}}e^{\theta_{j}}}{(e^{\theta_{i}}+e^{\theta_{j}})^{2}}\bm{v}(\bm{e}_{i}-\bm{e}_{j})(\bm{e}_{i}-\bm{e}_{j})^{\top}\bm{v}\\
 & \ge\sum_{(i,j):i>j}w_{ij}\frac{e^{-R}e^{R}}{(e^{-R}+e^{R})^{2}}\bm{v}(\bm{e}_{i}-\bm{e}_{j})(\bm{e}_{i}-\bm{e}_{j})^{\top}\bm{v}\\
 & \ge\frac{1}{4e^{2R}}\lambda_{n-1}(\bm{L}_{w}).
\end{align*}
Here the last inequality follows from the definition of $\lambda_{n-1}(\bm{L}_w)$. We also have that 
\begin{align*}
    \lambda_{n-1}(\bm{L}_{wz}^{\dagger}) = \|\bm{L}_{wz}\|^{-1}>0.
\end{align*}
Now consider the precondition gradient descent update 
\[
\bm{\theta}^{t+1}=\bm{\theta}^{t}-\eta\bm{L}_{wz}^{\dagger}\nabla\mathcal{L}(\bm{\theta}^{t}).
\]
Expanding the gradient difference we have 
\begin{align*}
\nabla\mathcal{L}(\bm{\theta}^{t+1}) & =\nabla \mathcal{L}(\bm{\theta}^{t})+\int_0^1 \nabla^{2}\mathcal{L}(\widetilde{\bm{\theta}}(\alpha))(\bm{\theta}^{t+1}-\bm{\theta}^{t}) \mathrm{d}\alpha\\
 & =\nabla \mathcal{L}(\bm{\theta}^{t})-\eta\left(\int_0^1 \mathcal{L}(\widetilde{\bm{\theta}}(\alpha))\mathrm{d}\alpha\right) \bm{L}_{wz}^{\dagger}\nabla\mathcal{L}(\bm{\theta}^{t})\\
 & =\left(\bm{I}-\eta\left(\int_0^1 \mathcal{L}(\widetilde{\bm{\theta}}(\alpha))\mathrm{d}\alpha\right)\bm{L}_{wz}^{\dagger}\right)\nabla\mathcal{L}(\bm{\theta}^{t}),
\end{align*}
where $\widetilde{\bm{\theta}}(\alpha) \coloneqq (1-\alpha)\bm{\theta}^t + \alpha \bm{\theta}^{t+1}$. Since $\bm{1}_n^{\top}\nabla\mathcal{L}(\bm{\theta}^{t})=0$ and $\bm{1}_n^{\top}\bm{L}_{wz}^{\dagger}\nabla\mathcal{L}(\bm{\theta}^{t})=0$,
\begin{align*}
\|\nabla\mathcal{L}(\bm{\theta}^{t+1})\| & \le\left[1-\frac{\eta}{4e^{2R}}\lambda_{n-1}(\bm{L}_{w})\lambda_{n-1}(\bm{L}_{wz}^{\dagger})\right]\|\nabla\mathcal{L}(\bm{\theta}^{t})\|.
\end{align*}
Then for sufficiently small $\eta>0$, $\|\nabla\mathcal{L}(\bm{\theta}^{t})\|$
goes to 0 as $t\rightarrow\infty$.

\paragraph*{Proof of Claim~\ref{claim:exist_unique_MLE}.}

Recall
\[
\mathcal{L}(\bm{\theta})=\frac{1}{L}\sum_{i,j:i>j}\sum_{l=1}^{L}w_{ij}\left(-y_{ji}^{(l)}(\theta_{i}-\theta_{j})+\log(1+e^{\theta_{i}-\theta_{j}})\right).
\]
Observe that since $0\le y_{ji}^{(l)}\le 1$, \begin{equation}
    -y_{ji}^{(l)}(\theta_{i}-\theta_{j})+\log(1+e^{\theta_{i}-\theta_{j}})\ge0\label{eq:nonneg}
\end{equation}
for each $i,j$. Since $|\theta_{i}^{\star}-\theta_{j}^{\star}|\le \log\kappa$ for any $(i,j)\in [n]^2$, 
\begin{align}
\mathcal{L}(\bm{\theta}^{\star}) & =\frac{1}{L}\sum_{i,j:i>j}\sum_{l=1}^{L}w_{ij}\left(-y_{ji}^{(l)}(\theta_{i}^{\star}-\theta_{j}^{\star})+\log(1+e^{\theta_{i}^{\star}-\theta_{j}^{\star}})\right)\nonumber\\
 & \le\sum_{i,j:i>j}w_{ij}\left(\log(\kappa)+\log(1+\kappa)\right)\nonumber\\
 & \le\sum_{i,j:i>j}w_{ij}2\log(2\kappa).\label{eq:L_theta_star}
\end{align}
Let $R$ be a large enough scalar such that
\begin{equation}
    \frac{R}{nL}\min_{\{i,j:w_{ij}>0\}}w_{ij}>\sum_{\{i,j:i>j\}}w_{ij}2\log(2\kappa).\label{eq:R}
\end{equation}
Now consider any $\bm{\theta}$ with $\|\bm{\theta}\|_{\infty}>R$.
Assume without loss of generality $\theta_{1}\ge\theta_{2}\ge\cdots\ge\theta_{n}$. Since we only consider
the case when $\bm{\theta}^{\top}\bm{1}_{n}=0$, $\theta_{1}-\theta_{n}>R-0$
and there exists some $k\le n-1$ such that $\theta_{k}-\theta_{k+1}>R/n$.
We then have a natural partition by two non-empty vertex set $\mathcal{S}_{1}=\{i:i>k\},\mathcal{S}_{2}=\{j:j\le k\}$.
By assumption for some $i>k,j\le k$, $w_{ij}>0$ and there is some
$l$ such that $y_{ji}^{(l)}=1$. Then combine this with (\ref{eq:nonneg}),
\begin{align}
\mathcal{L}(\bm{\theta}) & \ge\frac{1}{L}w_{ij}\left(-y_{ji}^{(l)}(\theta_{i}-\theta_{j})+\log(1+e^{\theta_{i}-\theta_{j}})\right)\nonumber\\
 & \ge\frac{1}{L}w_{ij}\left(R/n+\log(1)\right)\nonumber\\
 & \ge\frac{R}{nL}\min_{i,j:w_{ij}>0}w_{ij}\label{eq:L_theta}.
\end{align}
Putting (\ref{eq:L_theta_star}), (\ref{eq:R}), and (\ref{eq:L_theta}) together, we see that $\mathcal{L}(\bm{\theta}^{\star})\le\mathcal{L}(\bm{\theta})$
for any $\bm{\theta}$ such that $\bm{\theta}^{\top}\bm{1}_{n}=0$ and $\|\bm{\theta}\|_{\infty}>R$.
Then by the continuity of $\mathcal{L}$ there must exist a minimizer in
the closed and bounded set $\{\bm{\theta}:\|\bm{\theta}\|_{\infty}\le R:\bm{\theta}^{\top}\bm{1}_{n}=0\}$.
The restricted uniqueness is guaranteed by the restricted strong convexity
shown earlier, that is for any unit vector $\bm{v}\in\mathbb{R}^{n}$
such that $\bm{v}^{\top}\bm{1}_{n}=0$ and $\|\bm{\theta}\|_\infty \le R$, 
\[
\bm{v}^{\top}\nabla^{2}\mathcal{L}(\bm{\theta})\bm{v}\ge\frac{1}{4e^{2R}}\lambda_{n-1}(\bm{L}_{w}) >0.
\]

\section{Proof of Lemma~\ref{lemma:cr_conductance} \label{subsec:Proof_cr_conductance}}

This proof is mostly based on the proof of Lemma~28 in \cite{kelner2014almost},
which considers the case with integer weights. Let $\bm{v}=\bm{L}_{\mathcal{G}_{\widetilde{w}}}^{\dagger}(\bm{e}_{k}-\bm{e}_{l})$,
where $\bm{L}_{\mathcal{G}_{\widetilde{w}}}$ is the ${\widetilde{w}_{ij}}$-weighted graph Laplacian.
Then we can rewrite the LHS of (\ref{eq:congestion_conductance})
with 
\[
\sum_{(i,j):i>j}w_{ij}\left|(\bm{e}_{i}-\bm{e}_{j})^{\top}\bm{L}_{\mathcal{G}_{\widetilde{w}}}^{\dagger}(\bm{e}_{k}-\bm{e}_{l})\right|=\sum_{(i,j):i>j}w_{ij}\left|v_{i}-v_{j}\right|.
\]

For the rest of the proof we write $(i,j)$ as edges so $(i,j)$ and
$(j,i)$ are the same element. For any $c\in\mathbb{R}$ we define
two vertex sets:
\[
\mathcal{S}_{c}^{>}\coloneqq\{i\in\mathcal{V}:v_{i}>c\}\qquad\text{and}\qquad\mathcal{S}_{c}^{<}\coloneqq\{i\in\mathcal{V}:v_{i}<c\}.
\]

Recall that the volume of a vertex set $\mathcal{S}$ is defined as
\[
\mathrm{vol}(\mathcal{S})\coloneqq\sum_{(i,j)\in\mathcal{E}:i\in \mathcal S}w_{ij}.
\]
Let 
\[
\bar{c}\coloneqq\sup_{c}\left\{c:\mathrm{vol}(\mathcal{S}_{c}^{<})\le\frac{1}{2}\mathrm{vol}(\mathcal{V})\right\}
\]
 and 
\[
\underline{c}\coloneqq\inf_{c}\left\{c:\mathrm{vol}(\mathcal{S}_{c}^{>})\le\frac{1}{2}\mathrm{vol}(\mathcal{V})\right\}.
\]
For any $\epsilon>0$, 
\[
\mathrm{vol}(\mathcal{S}_{\bar{c}+\epsilon}^{>})\le \mathrm{vol}(\mathcal{V})-\mathrm{vol}(\mathcal{S}_{\bar{c}+\epsilon}^{<})\le \mathrm{vol}(\mathcal{V})/2.
\]
By the definition of $\underline{c}$, $\bar{c}+\epsilon\ge\underline{c}$.
This holds for every $\epsilon>0$ so $\bar{c}\ge\underline{c}$.
Fix some $c^{\star}\in [\underline{c}, \overline{c}]$, we have that
\begin{equation}
\mathrm{vol}(\mathcal{S}_{c^{\star}}^{>})\le1/2\qquad\text{and}\qquad\mathrm{vol}(\mathcal{S}_{c^{\star}}^{<})\le1/2.\label{eq:SCstar_vol}
\end{equation}

For any vertex set $\mathcal{S}\subset\mathcal{V}$, we denote $\partial\mathcal{S}\coloneqq\{(i,j):i\in \mathcal{S},v\in\mathcal{V}\setminus\mathcal{S}\}$.
We also define the flow and weight of the graph cut corresponding
to $\mathcal{S}$ as
\[
f(\mathcal{S})\coloneqq\sum_{(i,j)\in\partial\mathcal{S}}w_{ij}|v_{i}-v_{j}|\qquad\text{and}\qquad w(\mathcal{S})\coloneqq\sum_{(i,j)\in\partial\mathcal{S}}w_{ij}.
\]
Abusing the notation, for any set of edges $\mathcal{F}$ we write
$w(\mathcal{F})\coloneqq\sum_{(i,j)\in\mathcal{F}}w_{ij}$. Now consider
a sequence of real numbers $\{c_{i}\}$ defined recursively by $c_{0}\coloneqq c^{\star}$
and for any $t\ge0$,
\[
c_{t+1}\coloneqq c_{t}+\Delta_{t}\qquad\text{for}\qquad\Delta_{t}\coloneqq\frac{2f(\mathcal{C}_{t})}{w(\mathcal{C}_{t})},
\]
where to simplify the notation we also write $\mathcal{C}_{t}\coloneqq\mathcal{S}_{c_{t}}^{>}$.
At this point we will use the following fact. 

\begin{fact}For any vertex subset $\mathcal{S}\subset\mathcal{V}$,
\[
\sum_{(i,j)\in\partial\mathcal{S}}\widetilde{w}_{ij}(v_{i}-v_{j})\le1
\]

\end{fact}In a graph theoretical language, this fact says that the total flow going across any cut is at most the amount of flow going from source to sink. The exact proof is omitted as it requires the introduction
of a number of notions that are irrelevant to the rest of this paper.
Please see \cite{kelner2014almost} for the details. For any $t\ge0$,
from the definition of $\mathcal{C}_{t}$ we know that for any $i\in\mathcal{C}_{t}$
and $j\in\mathcal{V}\setminus\mathcal{C}_{t}$, $v_{i}\ge v_{j}$.
Then using the above fact we have 
\begin{align*}
f(\mathcal{C}_{t}) & =\sum_{(i,j):i\in\mathcal{C}_{t},v\in\mathcal{V}\setminus\mathcal{C}_{t}}w_{ij}|v_{i}-v_{j}|\\
 & =\sum_{(i,j):i\in\mathcal{C}_{t},v\in\mathcal{V}\setminus\mathcal{C}_{t}}w_{ij}(v_{i}-v_{j})\\
 & \le Z\sum_{(i,j):i\in\mathcal{C}_{t},v\in\mathcal{V}\setminus\mathcal{C}_{t}}\widetilde{w}_{ij}(v_{i}-v_{j})\le Z.
\end{align*}
Here $Z$ is a normalized factor defined as $Z\coloneqq\max_{ij}\{w_{ij}/\widetilde{w}_{ij}\}$.
We now show that $\mathrm{vol}(\mathcal{C}_{t})$ exponentially
decreases with $t$. Since $\mathcal{C}_{t+1}\subset\mathcal{C}_{t}$
for any $t\ge0$,
\begin{align*}
\mathrm{vol}(\mathcal{C}_{t+1}) & \le\mathrm{vol}(\mathcal{C}_{t})-w(\partial\mathcal{C}_{t}\setminus\partial\mathcal{C}_{t+1})\\
 & =\mathrm{vol}(\mathcal{C}_{t})-w(\partial\mathcal{C}_{t})+w(\partial\mathcal{C}_{t}\cap\partial\mathcal{C}_{t+1}).
\end{align*}
For any $(i,j)\in\partial\mathcal{C}_{t}\cap\partial\mathcal{C}_{t+1}$,
$i\in\mathcal{C}_{t+1}$ and $j\in\mathcal{V}\setminus\mathcal{C}_{t}$
(or the other way around). Then by the choice of $\Delta_{t}$,
\begin{align*}
f(\mathcal{C}_{t}) & =\sum_{(i,j)\in\partial\mathcal{C}_{t}}w_{ij}|v_{i}-v_{j}|\\
 & \ge\sum_{(i,j)\in\partial\mathcal{C}_{t}\cap\partial\mathcal{C}_{t+1}}w_{ij}|v_{i}-v_{j}|\\
 & \ge\sum_{(i,j)\in\partial\mathcal{C}_{t}\cap\partial\mathcal{C}_{t+1}}w_{ij}|c_{t+1}-c_{t}|\\
 & =w(\partial\mathcal{C}_{t}\cap\partial\mathcal{C}_{t+1})\cdot\Delta_{t}\\
 & =w(\partial\mathcal{C}_{t}\cap\partial\mathcal{C}_{t+1})\cdot\frac{2f(\mathcal{C}_{t})}{w(\mathcal{C}_{t})}.
\end{align*}
Therefore $w(\partial\mathcal{C}_{t}\cap\partial\mathcal{C}_{t+1})\le w(\mathcal{C}_{t})/2$
and 
\begin{align*}
\mathrm{vol}(\mathcal{C}_{t+1}) & \le\mathrm{vol}(\mathcal{C}_{t})-w(\partial\mathcal{C}_{t})+w(\partial\mathcal{C}_{t}\cap\partial\mathcal{C}_{t+1})\\
 & \le\mathrm{vol}(\mathcal{C}_{t})-\frac{w(\mathcal{C}_{t})}{2}\\
 & \le\mathrm{vol}(\mathcal{C}_{t})-\frac{1}{2}\mathrm{vol}(\mathcal{C}_{t})\Phi(\mathcal{G}_{w}),
\end{align*}
where the last inequality follows from the definition of the conductance
$\Phi(\mathcal{G}_{w})$. Now applying this recursively and use (\ref{eq:SCstar_vol}),
\[
\mathrm{vol}(\mathcal{C}_{t})\le\left(1-\frac{1}{2}\Phi(\mathcal{G}_{w})\right)^{t}\mathrm{vol}(\mathcal{C}_{0})\le\frac{1}{2}\left(1-\frac{1}{2}\Phi(\mathcal{G}_{w})\right)^{t}\mathrm{vol}(\mathcal{V}).
\]
As $\sum_{i\in\mathcal{S},j\in\mathcal{V}\setminus\mathcal{S}}w_{ij}\le\mathrm{vol}(\mathcal{S})$
for any $\mathcal{S}\subset\mathcal{V}$,
\[
\Phi(\mathcal{G}_{w})\coloneqq\min_{\mathcal{S}\subset\mathcal{V}}\frac{\sum_{i\in\mathcal{S},j\in\mathcal{V}\setminus\mathcal{S}}w_{ij}}{\min\{\mathrm{vol}(\mathcal{S}),\mathrm{vol}(\mathcal{V}\setminus\mathcal{S})\}}\le1.
\]
So $\mathrm{vol}(\mathcal{C}_{t})$ is decreasing in $t$ and for
any $t>2\log(\mathrm{vol}(\mathcal{V}))/\Phi(\mathcal{G}_{w})$, 
\begin{align*}
\mathrm{vol}(\mathcal{C}_{t}) & \le\frac{1}{2}\left(1-\frac{1}{2}\Phi(\mathcal{G}_{w})\right)^{t}\mathrm{vol}(\mathcal{V})\\
 & \le\frac{1}{2}e^{-\frac{1}{2}t\Phi(\mathcal{G}_{w})}\mathrm{vol}(\mathcal{V})\\
 & <\frac{1}{2}.
\end{align*}
By assumption, for any vertex $i\in\mathcal{V}$, $\mathrm{vol}(\{i\})\ge d_{\min}\ge 1$, so $\mathrm{vol}(\mathcal{C}_{t})<1/2$ implies $\mathrm{vol}(\mathcal{C}_{t})=\varnothing$.
Now let $r$ be the smallest integer such that $\mathrm{vol}(\mathcal{C}_{r+1})=\varnothing$,
we have $r\le2\log(\mathrm{vol}(\mathcal{V}))/\Phi(\mathcal{G}_{w})$.
Let $d_{i}\coloneqq\sum_{j:j\neq i}w_{ij}$. Then 
\begin{align*}
\sum_{i\in\mathcal{S}_{c^{\star}}^{>}}d_{i}|v_{i}-c^{\star}| & =\sum_{t=0}^{r}\sum_{i\in\mathcal{C}_{t}\setminus\mathcal{C}_{t+1}}d_{i}(v_{i}-c^{\star})\\
 & \overset{\text{(i)}}{\le}\sum_{t=0}^{r}\sum_{i\in\mathcal{C}_{t}\setminus\mathcal{C}_{t+1}}d_{i}(c_{t+1}-c^{\star})\\
 & \overset{\text{(ii)}}{\le}\sum_{t=0}^{r}\left(\mathrm{vol}(\mathcal{C}_{t})-\mathrm{vol}(\mathcal{C}_{t+1})\right)\sum_{s=0}^{t}\Delta_{s}\\
 & \overset{\text{(iii)}}{\le}\sum_{t=0}^{r}\mathrm{vol}(\mathcal{C}_{t})\Delta_{t}.
\end{align*}
Here (i) comes from the definition of $\mathcal{C}_{t+1}$, (ii)
comes from the definition of $d_{i}$ and choice of $c_{t+1}$, and
(iii) comes from rearrangement. Now we control the term $\mathrm{vol}(\mathcal{C}_{t})\Delta_{t}$
for any $t\ge0$. From the definition of $\Phi(\mathcal{G}_{w})$ we have
that 
\[
\Phi(\mathcal{G}_{w})\le\frac{w(\mathcal{C}_{t})}{\min\{\mathrm{vol}(\mathcal{C}_{t}),\mathrm{vol}(\mathcal{V})-\mathrm{vol}(\mathcal{C}_{t})\}}\le\frac{w(\mathcal{C}_{t})}{\mathrm{vol}(\mathcal{C}_{t})}
\]
where the last inequality follows from (\ref{eq:SCstar_vol}). Moreover
we established earlier in this proof that $f(\mathcal{C}_{t})\le Z$,
then
\[
\mathrm{vol}(\mathcal{C}_{t})\Delta_{t}\le\frac{w(\mathcal{C}_{t})}{\Phi(\mathcal{G}_{w})}\cdot\frac{2Z}{w(\mathcal{C}_{t})}=\frac{2Z}{\Phi(\mathcal{G}_{w})}.
\]
Therefore
\[
\sum_{i\in\mathcal{S}_{c^{\star}}^{>}}d_{i}|v_{i}-c^{\star}|\le\frac{2Zr}{\Phi(\mathcal{G}_{w})}.
\]
Similarly we can achieve 
\[
\sum_{i\in\mathcal{S}_{c^{\star}}^{<}}d_{i}|v_{i}-c^{\star}|\le\frac{2Zr}{\Phi(\mathcal{G}_{w})}.
\]
Finally, 
\begin{align*}
\sum_{(i,j):i>j}w_{ij}\left|v_{i}-v_{j}\right| & \le\sum_{(i,j):i>j}w_{ij}\left(\left|v_{i}-c^{\star}\right|+\left|v_{j}-c^{\star}\right|\right)\\
 & \le\sum_{i\in\mathcal{V}}d_{i}|v_{i}-c^{\star}|\\
 & \le\sum_{i\in\mathcal{S}_{c^{\star}}^{>}}d_{i}|v_{i}-c^{\star}|+\sum_{i\in\mathcal{S}_{c^{\star}}^{<}}d_{i}|v_{i}-c^{\star}|+\sum_{i:v_{i}=c^{\star}}d_{i}|v_{i}-c^{\star}|\\
 & \le\frac{4Zr}{\Phi(\mathcal{G}_{w})}\le\frac{8Z\log(\mathrm{vol}(\mathcal{V}))}{\Phi(\mathcal{G}_{w})^{2}}.
\end{align*}
This finishes the proof.

\section{Proofs for SDP-based reweighting} \label{app:sdp-proofs}
\subsection{Proofs for oracles in Section~\ref{sec:sdp}}
\begin{proof}[Proof of Theorem~\ref{thm:LP-oracle}]
Let $\bm{c} \in \R^{\cE_{\mathrm{SR}}}$ be defined by $c_{ij} = \langle \bm{L}_{ij}, \bm{X} \rangle.$ Then, we are interested in approximately solving the linear program (LP) $\min_{\bm{w} \in \cF} \langle \bm{c} , \bm{w} \rangle.$ Because $\bm{c} \geq 0$ and the set $\cF$ consists of the intersection of the positive quadrant with entrywise non-negative linear upper bounds, this LP is an instance of a packing LP. This class of programs can be approximately solved by specialized first-order algorithms.
In particular, the solver of~\cite{packing} yields a $(1-\epsilon)$ multiplicative approximation in time $\tilde{O}\left(\nicefrac{\mathrm{nnz}+n}{\epsilon}\right)$, where $\mathrm{nnz}$ is the number of non-zero entries in the matrix defining the constraints. By the construction of $\cF$, the sparsity of the corresponding constraints is simply $|\cE_{\mathrm{SR}}|.$ This complete the proof of the theorem. 
\end{proof}
\begin{proof}[Proof of Theorem~\ref{thm:greedy-oracle}]
Let $\bm{c} \in \R^{\cE_{\mathrm{SR}}}$ be defined by $c_{ij} \defeq \langle \bm{L}_{ij}, \bm{X} \rangle.$ Let $d \defeq 2pn$. Without loss of generality, we can assume that $d$ is an integer. Then, we are interested in approximately solving the following linear program (LP):
\begin{align*}
\max  \langle \bm{c} , \bm{w} \rangle &\\
\forall i \in V,\; \sum_{ij \in \cE_{\mathrm{SR}}} \bm{w}_{ij} &\leq d,\\
\forall ij \in \cE_{\mathrm{SR}}, \; \bm{0} \leq \bm{w}_{ij} &\leq 1.
\end{align*}
Notice that this is a relaxation of the maximum weight $d$-matching problem with weight $\bm{c}$ over the graph $\cG_{\mathrm{SR}}$. We are going to exploit this connection by showing that a greedy maximal-weight $d$-matching yields a \nicefrac{1}{2}-approximation to the optimum of this LP. To bound the value of this optimum, we will rely on the following dual LP:
\begin{align*}
\min  d\cdot \sum_{i \in V} s_i + \sum_{ij \in \cE_{\mathrm{SR}}} \ell_{ij} &\\
\forall ij \in \cE_{\mathrm{SR}}, s_i + s_j + \ell_{ij} &\geq \bm{c}_{ij},\\
\forall ij \in \cE_{\mathrm{SR}}, \ell_{ij} \geq 0,\\
\forall i \in V, s_i \geq 0
\end{align*}
The greedy $d$-matching is constructed in the following  natural way. First all edges are sorted in decreasing order of weight $\bm{c}.$ Then, edges are added to the matching in this order as long as their addition does not cause a degree constraint to become violated. Let $M$ be the resulting matching. We are now going to construct a feasible dual solution based on $M$. For each vertex $i \in V,$ let $s_i$ be the weight of the last matching edge incident to $i$ that was added to $M$. Notice that some $s_i$ may be 0 if all edges were exhausted before the degree constraint was violated. For any edge $ij \in M,$ let 
\begin{equation}\label{eq:dual-fitting}
\ell_{ij} =  \max\{\bm{c}_{ij} - s_i - s_j, 0\}.
\end{equation}
If $ij \notin M,$ simply let $\ell_{ij} = 0.$

We will now argue that the resulting dual solution is feasible. It suffices to show that all constraints $s_i + s_j + \ell_{ij} \geq \bm{c}_{ij}$ are satisfied. If $ij \in M,$ this is an immediate consequence of Equation~\ref{eq:dual-fitting}. If $ij \notin M,$ the addition of $ij$ must cause either $i$ or $j$ to violate the degree constraint. Suppose wlog that this was the case for vertex $i.$ Then, it must be the case that  $\bm{c}_{ij} \leq s_i,$ as edge $ij$ was considered after the last matching edge was added to $i.$ This proves that our dual solution is feasible. 

We now compare the value of the dual solution to that of the greedy matching. 
By construction, for each vertex $i,$ we have:
\begin{eqnarray*}
\sum_{j : ij \in M} \bm{c}_{ij} = \sum_{j : ij \in M \land \ell_{ij} = 0} \bm{c}_{ij} +\sum_{j : ij \in M \land \ell_{ij} >  0}\bm{c}_{ij} \geq\\ \sum_{j : ij \in M \land \ell_{ij} = 0} s_i + \sum_{j : ij \in M \land \ell_{ij} >  0} s_i + \nicefrac{\ell_{ij}}{2} = d \cdot s_i + \frac{1}{2} \sum_{j : ij \in M} \ell_{ij},
\end{eqnarray*}
where the last equality uses the fact that $s_i = 0$ if the degree of $i$ in M is less than $d.$ 
Summing over all vertices yields:
$$
2 \cdot \sum_{ij \in M} \bm{c}_{ij} = \sum_{i \in V}\sum_{j : ij \in M} \bm{c}_{ij} = d \cdot \sum_{i \in V} + \sum_{ij \in M} \ell_{ij}.
$$ 
As the right hand side is the dual value of our solution, we have shown that the greedy matching achieves a primal value that is within a factor of two of the optimal, as required.
\end{proof}

\subsection{Proofs for the analysis of Algorithms~\ref{alg:MMWU}}
\label{sec:sdp.apdx}

In this section, we verify that our use of Johnson--Lindenstrauss preserves the $\ell_2^2$-distances up to an $\epsilon$-multiplicative factor by proving Lemma~\ref{lem:jl}. We also demonstrate Lemma~\ref{lem:mmwu-rt} thereby justifying that Algorithm~\ref{alg:MMWU} runs in time nearly-linear with respect to the size of $\cG_{\mathrm{SR}}$.

Beginning with Lemma~\ref{lem:jl}, we first recall the statement of Johnson--Lindenstrauss provided by~\cite{achlioptasDatabasefriendlyRandomProjections2003}. The following can be recovered by setting $\beta = 8$, and observing that $\nicefrac{\epsilon^2}{2} - \nicefrac{\epsilon^2}{3} \geq \nicefrac{\epsilon^2}{6}$ for all $\epsilon \in [0, 1]$.

\begin{theorem}[Theorem 1.1 in~\cite{achlioptasDatabasefriendlyRandomProjections2003}]
\label{thm:jl}
Suppose $\mathbf{A} \in \R^{n \times d}$ and $\epsilon \in [0, 1]$ are given. For any $k \geq \nicefrac{120 \cdot \log n}{\epsilon^2}$, let $\mathbf{R} = \{ \pm \nicefrac{1}{\sqrt{k}} \}^{n \times k}$ be a random matrix with entries sampled independently and uniformly at random, and set $\mathbf{V} = \mathbf{A} \mathbf{R}$. If $\bm{a}_i \in \R^d$ and $\bm{v}_i \in \R^k$ denote the $i$-th row of $\mathbf{A}$ and $\mathbf{V}$ respectively, then
\begin{equation*}
(1 - \epsilon) \cdot \lVert \bm{a}_i - \bm{a}_j \rVert^2
\leq \lVert \bm{v}_i - \bm{v}_j \rVert^2
\leq (1 + \epsilon) \cdot \lVert \bm{v}_i - \bm{v}_j \rVert^2
\end{equation*}
\end{theorem}

We will also require a result which approximately computes the action of a matrix exponential where the matrix is Symmetric and Diagonally Dominant (SDD). Recall that a matrix $\bm{A} \in \R^{n \times n}$ is SDD if it is both symmetric, and for all $i \in [n]$, the entries satisfy $A_{ii} \geq \sum_{j \neq i} \lvert A_{ij} \rvert$. In~\cite{orecchiaApproximatingExponentialLanczos2012}, they apply the Lanczos method to obtain the following guarantee.

\begin{theorem}[Theorem 1.2 in~\cite{orecchiaApproximatingExponentialLanczos2012}]
\label{thm:expm}
Given an SDD matrix $\bm{A} \in \R^{n \times n}$, a vector $\bm{v} \in \R^n$, and $\delta > 0$, there exists an algorithm $\mathcal{A}$ which outputs a vector $\bm{u} \in \R^n$ satisfying
\begin{equation*}
\big\lVert \exp \{ - \bm{A} \} \cdot \bm{v} - \bm{u} \big\rVert_2
\leq \delta \cdot \lVert \bm{v} \rVert_2
\end{equation*}
in time $\tilde{O} \big( (\mathsf{nnz}(\bm{A}) + n) \cdot \log (2 + \lVert \bm{A} \rVert ) \big)$. Here, $\mathsf{nnz}(\bm{A})$ denotes the number of non-zero entries of $\bm{A}$, $\lVert \bm{A} \rVert$ denotes its spectral norm, and $\tilde{O}( \cdot )$ hides factors of $\mathrm{poly}(\log n)$ and $\mathrm{poly}( \log \nicefrac{1}{\delta})$.
\end{theorem}

We can now prove Lemma~\ref{lem:jl}.

\begin{proof}[Proof of Lemma~\ref{lem:jl}]
To show that $\tilde{\mathbf{X}}^{(t)} \in \Delta$, note that $\tilde{\mathbf{X}}^{(t)} \succeq \mathbf{0}$ follows immediately from $\tilde{\mathbf{X}}^{(t)}$ being the Gram matrix of $\mathbf{V}^{(t)}$. To check that its trace with $\Pi_{\perp \bm{1}}$ is unit, we can compute
\begin{equation*}
\big\langle \Pi_{\perp \bm{1}}, \tilde{\mathbf{X}}^{(t)} \big\rangle
= \big\langle \Pi_{\perp \bm{1}}, \mathbf{V}^{(t)} (\mathbf{V}^{(t)})^{\top} \big\rangle
= \frac{\langle \Pi_{\perp \bm{1}}, \mathbf{U}^{(t)} (\mathbf{U}^{(t)})^{\top} \rangle}{\langle \Pi_{\perp \bm{1}}, \mathbf{U}^{(t)} (\mathbf{U}^{(t)})^{\top} \rangle}
= 1
\end{equation*}

We then show that $\langle \mathbf{L}_{ij}, \tilde{\mathbf{X}}^{(t)} \rangle$ is an $\epsilon$-multiplicative factor of $\langle \mathbf{L}_{ij}, \mathbf{X}^{(t)} \rangle$. Let $\mathbf{A} \in \R^{n \times d}$ be the Cholesky factorization of the matrix exponential $\mathbf{W}^{(t)} = \exp \left\{-\eta \sum_{ij \in \cE_{\mathrm{SR}}} {w}^{(t)}_{ij} \bm{L}_{ij} \right\}$, i.e. $\mathbf{W}^{(t)} = \bm{A} \bm{A}^{\top}$. Denote $\bm{a}_i \in \R^d$ and $\bm{v}_i \in \R^k$ by the $i$-th row of $\mathbf{A}$ and $\mathbf{V}$ respectively. Because Algorithm~\ref{alg:MMWU} sets
\begin{equation*}
\bm{U}^{(t)}
= \exp \bigg\{-\eta \sum_{ij \in \cE_{\mathrm{SR}}} {w}^{(t)}_{ij} \bm{L}_{ij} \bigg\} \bm{R} \, ,
\end{equation*}
and $k$ is chosen so that $k = O \big( \nicefrac{\log n}{\epsilon^2} \big)$, Johnson--Lindenstrauss in Theorem~\ref{thm:jl} implies
\begin{equation*}
(1 - \epsilon) \cdot \lVert \bm{a}_i - \bm{a}_j \rVert^2
\leq \lVert \bm{v}_i - \bm{v}_j \rVert^2
\leq (1 + \epsilon) \cdot \lVert \bm{a}_i - \bm{a}_j \rVert^2
\end{equation*}
for all $(i, j) \in \mathcal{V} \times \mathcal{V}$. We use this to claim the following bound
\begin{equation}
\label{eqn:jl-1}
(1 - \epsilon) \cdot \big\langle \Pi_{\perp \bm{1}}, \bm{A}\bm{A}^{\top} \big\rangle
\leq \big\langle \Pi_{\perp \bm{1}}, \bm{U}^{(t)} (\bm{U}^{(t)})^{\top} \big\rangle
\leq (1 + \epsilon) \cdot \big\langle \Pi_{\perp \bm{1}}, \bm{A}\bm{A}^{\top} \big\rangle \, .
\end{equation}
One can see the RHS by applying Johnson--Lindenstrauss in this manner:
\begin{equation*}
\big\langle \Pi_{\perp \bm{1}}, \bm{U}^{(t)} (\bm{U}^{(t)})^{\top} \big\rangle
= \frac{1}{n} \cdot \sum_{i < j} \lVert \bm{u}_i - \bm{u}_j \rVert^2
\leq (1 +  \epsilon) \cdot \frac{1}{n} \cdot \sum_{i < j} \lVert \bm{a}_i - \bm{a}_j \rVert^2
= (1 +  \epsilon) \cdot \big\langle \Pi_{\perp \bm{1}}, \bm{A}\bm{A}^{\top} \big\rangle \, .
\end{equation*}
The LHS of inequality~\eqref{eqn:jl-1} then follows by applying the Johnson--Lindenstrauss lower bound.

Now, to finally show that $\langle \mathbf{L}_{ij}, \tilde{\mathbf{X}}^{(t)} \rangle$ is an $\epsilon$-multiplicative factor of $\langle \mathbf{L}_{ij}, \mathbf{X}^{(t)} \rangle$, it suffices to demonstrate
\begin{equation}
\label{eqn:jl-2}
\big( 1 - O(\epsilon) \big) \cdot \langle \mathbf{L}_{ij}, \mathbf{X}^{(t)} \rangle
\leq \langle \mathbf{L}_{ij}, \tilde{\mathbf{X}}^{(t)} \rangle
\leq \big( 1 + O(\epsilon) \big) \cdot \langle \mathbf{L}_{ij}, \mathbf{X}^{(t)} \rangle \, .
\end{equation}
Because Algorithm~\ref{alg:MMWU} sets $\bm{V}^{(t)} = \nicefrac{\bm{U}^{(t)}}{\sqrt{\langle \Pi_{\perp \bm{1}}, \bm{U}^{(t)} (\bm{U}^{(t)})^T\rangle}}$, we have 
\begin{equation*}
\langle \mathbf{L}_{ij}, \tilde{\mathbf{X}}^{(t)} \rangle
= \big\langle \mathbf{L}_{ij}, \mathbf{V}^{(t)} (\mathbf{V}^{(t)})^{\top} \big\rangle
= \frac{\langle \mathbf{L}_{ij}, \mathbf{U}^{(t)} (\mathbf{U}^{(t)})^{\top} \rangle}{\langle \Pi_{\perp \bm{1}}, \mathbf{U}^{(t)} (\mathbf{U}^{(t)})^{\top} \rangle}
= \frac{\lVert \bm{u}_i - \bm{u}_{j} \rVert^2}{\langle \Pi_{\perp \bm{1}}, \mathbf{U}^{(t)} (\mathbf{U}^{(t)})^{\top} \rangle} \, .
\end{equation*}
On the other hand, setting $\mathbf{A}$ to be the Cholesky factorization of $\mathbf{W}^{(t)}$ determines
\begin{equation*}
\langle \mathbf{L}_{ij}, \mathbf{X}^{(t)} \rangle
= \frac{\langle \bm{L}_{ij}, \bm{W}^{(t)} \rangle}{\langle \Pi_{\perp \bm{1}}, \bm{W}^{(t)} \rangle}
= \frac{\langle \bm{L}_{ij}, \bm{A} \bm{A}^{\top} \rangle}{\langle \Pi_{\perp \bm{1}}, \bm{A} \bm{A}^{\top} \rangle}
= \frac{\lVert \bm{a}_i - \bm{a}_{j} \rVert^2}{\langle \Pi_{\perp \bm{1}}, \bm{A} \bm{A}^{\top} \rangle}
\end{equation*}
The RHS of inequality~\eqref{eqn:jl-2} then follows by applying the upper bound in Johnson--Lindenstrauss, and the lower bound in inequality~\eqref{eqn:jl-1} to get the following.
\begin{equation*}
\langle \mathbf{L}_{ij}, \tilde{\mathbf{X}}^{(t)} \rangle
= \frac{\lVert \bm{u}_i - \bm{u}_{j} \rVert^2}{\langle \Pi_{\perp \bm{1}}, \mathbf{U}^{(t)} (\mathbf{U}^{(t)})^{\top} \rangle}
\leq \frac{1 + \epsilon}{1 - \epsilon} \cdot \frac{\lVert \bm{a}_i - \bm{a}_{j} \rVert^2}{\langle \Pi_{\perp \bm{1}}, \mathbf{A} \mathbf{A}^{\top} \rangle}
\leq (1 + 2 \epsilon) \cdot \frac{\lVert \bm{a}_i - \bm{a}_{j} \rVert^2}{\langle \Pi_{\perp \bm{1}}, \mathbf{A} \mathbf{A}^{\top} \rangle}
= \big( 1 + O(\epsilon) \big) \cdot \langle \mathbf{L}_{ij}, \mathbf{X}^{(t)} \rangle
\end{equation*}
The LHS of inequality~\eqref{eqn:jl-2} can then be derived similarly
\begin{equation*}
\langle \mathbf{L}_{ij}, \tilde{\mathbf{X}}^{(t)} \rangle
= \frac{\lVert \bm{u}_i - \bm{u}_{j} \rVert^2}{\langle \Pi_{\perp \bm{1}}, \mathbf{U}^{(t)}}
\geq \frac{1 - \epsilon}{1 + \epsilon} \cdot \frac{\lVert \bm{a}_i - \bm{a}_{j} \rVert^2}{\langle \Pi_{\perp \bm{1}}, \mathbf{A} \mathbf{A}^{\top} \rangle}
\geq (1 - 2 \epsilon) \cdot \frac{\lVert \bm{a}_i - \bm{a}_{j} \rVert^2}{\langle \Pi_{\perp \bm{1}}, \mathbf{A} \mathbf{A}^{\top} \rangle}
= \big( 1 - O(\epsilon) \big) \cdot \langle \mathbf{L}_{ij}, \mathbf{X}^{(t)} \rangle
\end{equation*}
as required.
\end{proof}

Using Theorem~\ref{thm:expm}, we can also prove Lemma~\ref{lem:mmwu-rt}.

\begin{proof}[Proof of Lemma~\ref{lem:mmwu-rt}]
We bound the running time by considering each line of Algorithm~\ref{alg:MMWU}. As $k=O(\nicefrac{\log(n)}{\epsilon^2})$, Line~\ref{line:jl-matrix} only requires nearly-linear time. By Theorem 3.2 in~\cite{orecchiaApproximatingExponentialLanczos2012}, for any graph $\cG=(V,E)$, the action of the heat kernel $\exp\{-t \bm{L_{\cG}}\}$ for $t \geq 0$ can be in computed in nearly-linear time in the graph size. Line~\ref{line:jl-comp} requires $k$ such computations and hence also runs in nearly-linear time. The normalization of the embedding $\bm{U}^{(t)}$ requires computing the inner product $\langle \Pi_{\perp \bm{1}} ,\bm{U}^{(t)} (\bm{U}^{(t)})^T \rangle $. This can be achieved easily by noticing that:
$$
\langle \Pi_{\perp \bm{1}} ,\bm{U}^{(t)} (\bm{U}^{(t)})^T \rangle = \mathrm{Tr}(\bm{U}^{(t)} (\bm{U}^{(t)})^T ) - \frac{1}{n} \|\bm{U}^{(t)} \bm{1}\|^2.
$$
As the dimension of $U^{(t)}$ is $n \times k$, both of these terms can be computed in $O(nk)$ arithmetic operations, which is nearly-linear in $n.$ Similarly, the computation of each edge gain $\bm{c}_{ij}$ only requires computing the squared distance between the $i$ and $j$ column of $\bm{V}^{(t)}$, which can be achieved in time $O(k).$ Hence, all the gains can be computed in nearly-linear time. Finally, by Theorems~\ref{thm:LP-oracle} and~\ref{thm:greedy-oracle}, Line~\ref{line:oracle} also runs in nearly-linear time. Hence, every iteration of the main loop runs in nearly-linear time. As there are only $O(\nicefrac{\log n}{\epsilon^2})$ iterations, Algorithm~\ref{alg:MMWU} runs in nearly-linear time.
\end{proof}

\section{Sampling in clusters\label{sec:cluster}}

In this section we consider ranking in a sampling regime structured
with underlying clusters. Suppose that there is an underlying cluster
labeling $\varphi:[n]\rightarrow[k]$ such that item $i$ belongs
to cluster $\varphi(i)$. We assume each edge $(i,j)$ in the comparison graph $\mathcal{E}$ is drawn
independently with probability $p_{t}$ if $\varphi(i)=\varphi(j)=t$
for some $t\in[k]$ and probability $q$ if otherwise. Without loss of generality, we let the first $n_{1}$ items
to be cluster $1$, the next $n_{2}$ items to be cluster 2, etc. To distinguish
from the rest of the paper, we denote the sampled comparison graph
as $\mathcal{G}_{\mathrm{CL}}=(\mathcal{V}=[n],\mathcal{E}_{\mathrm{CL}})$
where $\mathrm{CL}$ stands for clusters.

We consider the case of $p_{t}>q$ for all $t$. It is easy to
see that this is a special case of semi-random sampling. The spectral profile of this setting can be vastly different from a Erd\H{o}s-R\'{e}nyi comparison graph. Thus a naive application of Theorem~\ref{thm:weight_MLE_spectral} would be unsatisfactory.

\begin{proposition}\label{prop:semi-random} Suppose that there is
an underlying cluster labeling $\varphi:[n]\rightarrow[k]$ and $(i,j)\in\mathcal{E}$
is drawn independently with probability $p_{t}$ if $\varphi(i)=\varphi(j)=t$
for some $t\in[k]$ and probability $q$ if otherwise. Let $\mathcal{G}=(\mathcal{V}=[n],\mathcal{E})$ be the comparison graph,
then there exists a subgraph of $\mathcal{G}$ that is an Erd\H{o}s-R\'{e}nyi
graph.

\end{proposition}\begin{proof}An equivalent way to write $\mathcal{E}$
is the following: let $\delta_{ij}\sim\mathrm{Unif}(0,1)$ be i.i.d.
random variables and 
\[
\mathcal{E}=\{(i,j):(\delta_{ij}\le p_{t}\text{ and }\varphi(i)=\varphi(j)=t\text{ for some }t\in[k])\text{ or }\delta_{ij}\le q\}.
\]
We also consider another graph $\mathcal{G}_{1}=(\mathcal{V},\mathcal{E}_{1})$
where $\mathcal{E}_{1}=\{(i,j):\delta_{ij}\le q\}$. It is easy to
see $\mathcal{G}_{1}$ is an Erd\H{o}s\textendash R\'enyi graph with uniform sampling
probability $q$ and also a subgraph of $\mathcal{G}$. \end{proof}

In addition to the assumptions in the semi-random setting, we assume
$n_{t}p_t\ge C\log (n)$ for some absolute constant $C>0$.  
We are interested in the vanilla, unweighted MLE, i.e. the solution of (\ref{eq:weighted_MLE}) with $w_{ij} = \mathds{1}_{(i,j)\in \mathcal{E}_{\mathrm{CL}}}$. The following result shows $\widehat{\bm{\theta}}$ achieves $\widetilde{O}(1/\sqrt{nqL})$ rate in $\ell_{\infty}$ error.

\begin{theorem}\label{thm:MLE_cluster}

Suppose that $nq\ge C_{1}\log (n)$ and $n_{t}p_{t}\ge C_{1}\log (n)$.
for some large enough constants $C_{1}>0$. Suppose
$nqL\ge C_{2}\kappa^{4}\log^3(n)$  and $n^{2}q^{2}L\ge C_{2}\kappa^{4}n_{t}p_{t}\log^3(n)$
for some large enough constant $C_{2}>0$. 
for any $t\in[n]$. Then with probability at least $1-O(n^{-10})$, for
any $(k,l)$, the unweighted MLE $\widehat{\bm{\theta}}$ satisfies
\[
\|\bm{\widehat{\bm{\theta}} - \theta}^\star  \|_\infty\le C_{3}\kappa\sqrt{\frac{\log(n)}{nqL}}
\]
for some constants $C_{3}>0$. As a result, the top-$K$ items are
recovered exactly as long as 
\[
n^2 qL\ge C_{4}\frac{\kappa^{2}n\log(n)}{\Delta_{K}^{2}}
\]
for some large enough constant $C_{5}$.

\end{theorem}The result of this theorem matches what we can
get in Theorem~\ref{thm:weighted_MLE_semirandom} using
reweighting, with a slight difference in the sample complexity requirement.
This suggests that at least in some non-uniform sampling models, reweighting
is unnecessary for MLE. 

\subsection{Proof of Theorem~\ref{thm:MLE_cluster} \label{subsec:Proof_cluster}}

We start by applying Lemma~\ref{thm:weighted_MLE} with $w_{ij} = \mathds{1}_{(i,j)\in \mathcal{E}_{\mathrm{CL}}}$. For any $k\neq l$, let $B_{kl}, Q_{kl}$ be some real number that we will specify later such that
\begin{align}
B_{kl} & \ge C_1\sqrt{\frac{\kappa}{L}\bm{\Omega}_{kl}(\bm{L}_{z})\log(n)};\label{eq:B_cluster}\\
Q_{kl} & \ge \sum_{(i,j)\in\mathcal{E}_{\mathrm{CL}}:i>j}B_{ij}^{2}\left|(\bm{e}_{k}-\bm{e}_{l})^{\top}\bm{L}_{z}^{\dagger}(\bm{e}_{i}-\bm{e}_{j})\right|\label{eq:Q_cluster}.
\end{align}
Here $C_1>0$ is some large enough constant and 
\[\bm{L}_z = \sum_{(i,j)\in \mathcal{E}_\mathrm{CL}:i>j} z_{ij}(\bm{e}_i -\bm{e}_j)(\bm{e}_i -\bm{e}_j)^\top .\]Suppose that  $Q_{kl}\le4B_{kl}$
for any $(k,l)$. By Lemma~\ref{thm:weighted_MLE}, with probability
at least $1-n^{-10}$, we have for any $(k,l)$, 
\[
\left|\left(\widehat{\theta}_{k}-\widehat{\theta}_{l}\right)-\left(\theta_{k}^{\star}-\theta_{l}^{\star}\right)\right|\le B_{kl}.
\]

We now let 
\[
{B}_{kl}\coloneqq C_2\sqrt{\frac{\kappa^{2}\log(n)}{(n_{t}p_{t} \vee nq)L}
}\]
 if $k,l$ are both in some cluster $t$ and 
\[
{B}_{kl}\coloneqq C_2\sqrt{\frac{\kappa^{2}\log(n)}{nqL}}\]
if $k,l$ are not in the same cluster. We also let 
\[Q_{kl}\coloneqq C_3\frac{ \kappa^{2}\log^{2}(n)}{nqL}\]
Here $C_2, C_3>0$ are some large enough constants.
constant. We denote $\mathcal{G}_{B^2}$ as the graph $\mathcal{G}_{\mathrm{CL}}$ equipped with weights $\{B_{ij}^2\}$. 

We now show (\ref{eq:B_cluster}) and (\ref{eq:Q_cluster}) hold using the same strategy in the proof of Theorem~\ref{thm:weight_MLE_spectral}.
First, we control effective resistance and graph conductance with the following lemmas. The proofs are deferred to Section~\ref{subsec:proof_effres_cluster} and \ref{subsec:proof_conduct_cluster}.

\begin{lemma}\label{lemma:effres_cluster} With probability at least
$1-O(n^{-10})$,
\[
\bm{\Omega}_{kl}(\bm{L}_{z})\le\frac{\kappa}{n_{t}p_{t} \vee nq}
\]
 if $k,l$ are both in some cluster $t$ and 
\[
\bm{\Omega}_{kl}(\bm{L}_{z})\le \frac{\kappa}{nq}
\]
if $k,l$ are not in the same cluster.

\end{lemma}

\begin{lemma} \label{lemma:B_conductance} Let $\mathcal{G}_{B^2}$
be $\mathcal{G}_{\mathrm{CL}}$ with weights $\{B_{ij}^2\}_{(i,j)\in\mathcal{E}_{\mathrm{CL}}}$. The graph conductance of $\mathcal{G}_{B^2}$ satisfies
\[
\Phi(\mathcal{G}_{B^2})\ge\frac{1}{8}.
\]

\end{lemma}
From Lemma~\ref{lemma:effres_cluster}, we can see that (\ref{eq:B_cluster}) is satisfied as long as $C_2$ is large enough. By Lemma~\ref{lemma:B_conductance} we have that
\begin{equation}
    \sum_{(i,j)\in\mathcal{E}_{\mathrm{CL}}:i>j}{B}_{ij}^2\left|(\bm{e}_{i}-\bm{e}_{j})^{\top}\bm{L}_{z}^{\dagger}(\bm{e}_{k}-\bm{e}_{l})\right|\le\max_{(i,j)\in\mathcal{E}_{\mathrm{CL}}}\left\{ {B}_{ij}^2\right\} \cdot\frac{8\log\left(\sum_{(i,j)\in\mathcal{E}_{\mathrm{CL}}:i>j}{B}_{ij}^2\right)}{\Phi(\mathcal{G}_{B^2})^{2}}.\label{eq:B_cluster_conduction}
\end{equation}
For all $(k,l)\in[n]^2$, 
\[{B}_{kl}^2\le C_2^2\frac{\kappa^{2}\log(n)}{nqL} \le 1\]
as long as $nqL\ge C_{4}\kappa^{4}\log^3(n)$ for some large enough constant $C_{4} > 0$. Then provided that $C_3$ is large enough, (\ref{eq:B_cluster_conduction}) becomes 
\begin{align*}
\sum_{(i,j)\in\mathcal{E}_{\mathrm{CL}}:i>j}{B}_{ij}^2\left|(\bm{e}_{i}-\bm{e}_{j})^{\top}\bm{L}_{z}^{\dagger}(\bm{e}_{k}-\bm{e}_{l})\right| & \le C_2^2\frac{\kappa^{2}\log(n)}{nqL}\cdot\frac{8\log(n^{2})}{(1/8)^{2}}\\
 & \le C_3\frac{\kappa^{2}\log^{2}(n)}{nqL} = Q_{kl}.
\end{align*}
This satisfies (\ref{eq:Q_cluster}). Using the assumption that $nqL\ge C_{4}\kappa^{4}\log^3(n)$
and $n^{2}q^{2}L\ge C_{5}\kappa^{4}n_{t}p_{t}\log^3(n)$ for any $t\in[n]$ with some
large enough constants $C_{4}, C_5 > 0$, we have that $Q_{kl}\le4B_{kl}$
for any $(k,l)\in[n]^2$. 

Now that we have checked all conditions for Lemma~\ref{thm:weighted_MLE}, we conclude that
\[
\left|\left(\widehat{\theta}_{k}-\widehat{\theta}_{l}\right)-\left(\theta_{k}^{\star}-\theta_{l}^{\star}\right)\right|\le B_{kl} = C_{2}\kappa\sqrt{\frac{\log(n)}{nqL}}.
\]

\subsection{Proof of Lemma~\ref{lemma:effres_cluster}\label{subsec:proof_effres_cluster}}

We consider two subgraphs $\mathcal{G}_{1}=(\mathcal{V},\mathcal{E}_{1})$
and $\mathcal{G}_{2}=(\mathcal{V},\mathcal{E}_{2})$ of $\mathcal{G}$,
where $\mathcal{E}_{1}$ consists of all edges within each cluster,
and $\mathcal{E}_{2}$ consists of the edges of an underlying Erd\H{o}s\textendash R\'enyi
subgraph as in Proposition~\ref{prop:semi-random}. Let their unweighted
graph Laplacian be $\bm{L}_{1}$ and $\bm{L}_{2}$. Let their $z$-weighted graph Laplacian be  $\bm{L}_{z, 1}$ and $\bm{L}_{z, 2}$. Combining Lemma~\ref{lem:B} and \ref{lemma:spectral_ER}, for $k, l$
both in some cluster $t$,
\[
\bm{\Omega}_{kl}(\bm{L}_{z,1})\le\frac{8\kappa}{\lambda_{n_{t}-1}(\bm{L}_{1})}\le \frac{16\kappa}{n_{t}p_t};
\]
while for any $k,l$ (including those in cluster $t$), we look at $\mathcal{G}_2$ to see
\[
\bm{\Omega}_{kl}(\bm{L}_{z,2})\le\frac{8\kappa}{\lambda_{n-1}(\bm{L}_{2})}\le \frac{16\kappa}{nq}.
\]
Combining this with Rayleigh's law of monotonicity (see Lemma~\ref{lemma:monotonicity}),
we reach the claimed result. 

\subsection{Proof of Lemma~\ref{lemma:B_conductance}\label{subsec:proof_conduct_cluster}}

From Lemma~\ref{lemma:conductance_eigen} we have
\begin{equation}
 \Phi(\mathcal{G}_{B^2})\ge\frac{1}{2}\lambda_{n-1}(\bm{D}_{B^2}^{-1/2}\bm{L}_{B^2}\bm{D}_{B^2}^{-1/2}).\label{eq:G_B2_conductance}
\end{equation}
Here $\bm{D}_{B^2}$ is the diagonal matrix with $[\bm{D}_{B^2}]_{ii}=\sum_{j:(i,j)\in\mathcal{E}}B^2_{ij}$
and
\begin{align*}
\bm{L}_{B^2} & =\sum_{(i,j)\in\mathcal{E}_{\mathrm{CL}},i>j}B^2_{ij}(\bm{e}_{i}-\bm{e}_{j})(\bm{e}_{i}-\bm{e}_{j})^{\top}\\
 & =\sum_{(i,j),i>j}\underbrace{B^2_{ij}\delta_{ij}(\bm{e}_{i}-\bm{e}_{j})(\bm{e}_{i}-\bm{e}_{j})^{\top}}_{\eqqcolon\bm{L}_{ij}},
\end{align*}
where $\delta_{ij}=\mathds{1}_{\{(i,j)\in\mathcal{E}_{\mathrm{CL}}\}}$.
To control the right hand side of (\ref{eq:G_B2_conductance}), we give a lower bound of $\lambda_{n-1}(\bm{L}_{B^2})$ and an upper bound of the diagonal entries of $\bm{D}_{B^2}$ in the following lemmas. The proofs are
deferred to the end of this section.
\begin{lemma}
    \label{lem:B2_eigengap } Instate the assumption of Lemma~\ref{lemma:B_conductance}. With probability at least
$1-O(n^{-10})$, \[\lambda_{n-1}(\bm{L}_{B^2})\ge \frac{C\kappa^{2}\log(n)}{2L},\]where $C$ is some constant.
\end{lemma}

\begin{lemma}\label{lemma:tildeB_degree} Instate the assumption of Lemma~\ref{lemma:B_conductance}. With probability at least
$1-O(n^{-10})$, 
\[
 \sum_{j;(i,j)\in \mathcal{E}}B^2_{ij}\le \frac{4C\kappa^{2}\log(n)}{L}
\]
for all $i\in[n]$. Here $C$ is a constant.

\end{lemma}
With these two lemmas, by Sylvester's law of inertia (Lemma~\ref{lemma:sylvester}),
\begin{align*}
\Phi(\mathcal{G}_{B^2}) & \ge\frac{1}{2}\lambda_{n-1}(\bm{D}_{B^2}^{-1/2}\bm{L}_{B^2}\bm{D}_{B^2}^{-1/2})\\
 & \ge\frac{1}{2}\lambda_{n-1}(\bm{L}_{B^2})\lambda_{n}(\bm{D}_{B^2}^{-1})\\
 & \ge \frac{1}{8}.
\end{align*}

\paragraph{Proof of Lemma~\ref{lem:B2_eigengap }.}
The proof follows similar strategy to Section 5.3.3 in \cite{tropp2015matrix}. Let $\bm{R}\in\mathbb{R}^{(n-1)\times n}$
be a partial isometry such that $\bm{R}\bm{R}^{\top}=\bm{I}_{n-1}$
and $\bm{R}\bm{1}_{n}=\bm{0}$. Then $\lambda_{n-1}(\bm{L}_{B^2})=\lambda_{n-1}(\bm{R}\bm{L}_{B^2}\bm{R}^{\top})$
and we will use matrix Chernoff to control the latter term. 

Recall that 
\[\bm{L}_{ij} = B^2_{ij}\delta_{ij}(\bm{e}_{i}-\bm{e}_{j})(\bm{e}_{i}-\bm{e}_{j})^{\top}.\]
For all
$(i,j)$ such that $i>j$, 
\[
0\le\lambda_{n-1}(\bm{R}\bm{L}_{ij}\bm{R}^{\top})\le\lambda_{1}(\bm{R}\bm{L}_{ij}\bm{R}^{\top})=\lambda_{1}(\bm{L}_{ij})\le 2C_2\frac{ \kappa^2\log(n)}{nqL}.
\]
Because $\mathbb{P}[L_{ij}^{t}=1\mid R_{ij}^{t}=1]\ge\frac{2\kappa}{(1+\kappa)^{2}}\ge1/(2\kappa),$
\begin{align*}
\lambda_{n-1}(\mathbb{E}\bm{R}\bm{L}_{B^2}\bm{R}^{\top}) & =\lambda_{n-1}\left(\bm{R}\sum_{t=1}^{n}\sum_{i>j}\mathbb{E}\bm{L}_{ij}\bm{R}^{\top}\right)\\
 & =\lambda_{n-1}\left(\bm{R}\sum_{t=1}^{n}\sum_{i>j}B^2_{ij}\mathbb{E}\delta_{ij}(\bm{e}_{i}-\bm{e}_{j})(\bm{e}_{i}-\bm{e}_{j})^{\top}\bm{R}^{\top}\right)
\end{align*}
When $i,j$ are in the same cluster $t$, \[B^2_{ij}\mathbb{E}\delta_{ij}\ge C_2\frac{ p_t\kappa^{2}\log(n)}{(n_{t}p_{t} \vee nq)L}\ge C_2\frac{\kappa^{2}\log(n)}{nL};\]
when $i,j$ are not in the same cluster, \[B^2_{ij}\mathbb{E}\delta_{ij}\ge C_2\frac{ q\kappa^{2}\log(n)}{nqL} =C_2\frac{\kappa^{2}\log(n)}{nL}.\]
Then
\begin{align*}
\lambda_{n-1}(\mathbb{E}\bm{R}\bm{L}_{B^2}\bm{R}^{\top}) & \ge C_2\frac{\kappa^{2}\log(n)}{nL}\cdot\lambda_{n-1}\left(\bm{R}\sum_{t=1}^{n}\sum_{i>j}(\bm{e}_{i}-\bm{e}_{j})(\bm{e}_{i}-\bm{e}_{j})^{\top}\bm{R}^{\top}\right)\\
 & =C_2\frac{\kappa^{2}\log(n)}{nL}\cdot\lambda_{n-1}\left[\bm{R}\left(n\bm{I}_{n}-\bm{1}_{n}\bm{1}_{n}^{\top}\right)\bm{R}^{\top}\right]\\
 & =C_2\frac{\kappa^{2}\log(n)}{nL}\cdot\lambda_{n-1}\left[n\bm{I}_{n-1}\right]\\
 & =C_2\frac{\kappa^{2}\log(n)}{L}.
\end{align*}
The second to last line holds since $\bm{R}\bm{R}^{\top}=\bm{I}_{n-1}$
and $\bm{R}\bm{1}_{n}=\bm{0}$. Then by matrix Chernoff (see e.g. Theorem~5.1.1 in \cite{tropp2015matrix}),
\begin{align*}
\mathbb{P}\left\{ [\lambda_{n-1}(\bm{R}\bm{L}_{B^2}\bm{R}^{\top})]\le \frac{C_2\kappa^{2} \log(n)}{2L}\right\}  & \le n\cdot\left[\frac{e^{-1/2}}{(1/2)^{1/2}}\right]^{nq/2}\\
 & \le n^{-10}
\end{align*}
as long as $nq\ge C_1\log (n)$ for some large enough constant $C_1$. The $nq/2$ term in the exponent comes from the fact that $\lambda_{n-1}(\mathbb{E}\bm{R}\bm{L}_{B^2}\bm{R}^{\top})/\lambda_{1}(\bm{L}_{ij})\ge nq/2$.
\paragraph{Proof of Lemma~\ref{lemma:tildeB_degree}.}
Let $\varphi(i)=t$. We split the summation with
\begin{align*}
\sum_{j:(i,j)\in\mathcal{E}}B^2_{ij} & =\sum_{j:(i,j)\in\mathcal{E},\varphi(j)=t}B^2_{ij}+\sum_{j:(i,j)\in\mathcal{E},\varphi(j)\neq t}B^2_{ij}\\
 & =\frac{C_1\kappa^{2}\log(n)}{L}\left[\sum_{j:(i,j)\in\mathcal{E},\varphi(j)=t}\frac{1}{n_{t}p_t \vee nq}+\sum_{j:(i,j)\in\mathcal{E},\varphi(j)\neq t}\frac{1}{nq}\right],
\end{align*}
where $C_1>0$ is a constant. 
By assumption for any $t\in [k]$, $nq\ge C_2 \log(n)$ and $n_{t}p_t\ge C_3\log (n)$, where $C_2, C_3>0$ are some constants. Applying standard Chernoff bound, we have that with probability at least $1-4n^{-10}$, for all $i\in[n]$,
\[
\frac{1}{2}n_{t}p_t\le\left|\{j:(i,j)\in\mathcal{E},\varphi(j)=t\}\right|\le2n_{t}p_t
\]
and 
\[
\frac{1}{2}nq\le\left|\{j:(i,j)\in\mathcal{E}\}\right|\le2nq
\]
as long as $C_2, C_3$ are large enough. Then 
\begin{align*}
\sum_{j:(i,j)\in\mathcal{E}}B^2_{ij} & \le C_1\frac{\kappa^{2}\log(n)}{L}\left[2n_{t}p_t \cdot \frac{1}{n_{t}p_t \vee nq}+2nq \cdot\frac{1}{nq}\right]\\
 & \le 4C_1\frac{\kappa^{2}\log(n)}{L}.
\end{align*}

\section{Experiment setup for Figure~\ref{fig:topK} } \label{app:experiments}

To check our result, we implement the Algorithm~\ref{alg:main} and apply it to some simulated data generated with semi-random sampling. Our semi-random graph is generated as follows:
\begin{enumerate}
    \item  Generate an Erd\H{o}s\textendash R\'enyi random graph $\mathcal{G}_{\mathrm{ER}}$ with $n$ vertices and edge probability $p$.
    \item Randomly select a subset $A$ of $n/3$ vertices in the first $n/2$ vertices, and a subset $B$ of $n/3$ vertices in the last $n/2$ vertices.
    \item Form $\mathcal{G}_{\mathrm{SR}}$ by adding all edges between vertices in $A$ and all edges between vertices in $B$ to $\mathcal{G}_{\mathrm{ER}}$.
\end{enumerate}
We then generate the comparison data following the BTL model with latent score $\bm{\theta}^\star$, where $\theta_i^\star = \Delta_K$ for $i = 1,\ldots, K$ and $\theta_i^\star = 0$ for $i = K+1, \ldots, n$. Similar to the example in Figure~\ref{fig:semirandom}, this quickly ruins the nice spectral properties observed in uniform sampling. 

We choose the parameters to be $n = 200, K = 10, L = 10, p = 0.25$, and $\Delta_K$ varying from 0.02 to 0.62. For each set of parameters, we do 50 independent trials. In each trial, we compute two MLE estimates of ${\bm{\theta}}^\star$ and the corresponding top-$K$ items. The first one is the vanilla MLE using all data on $\mathcal{G}_{\mathrm{ER}}$ and the second one is the weighted MLE given by Algorithm~\ref{alg:main}. We compare the top-$K$ recovery accuracy, i.e. the proportion of top-$K$ items that are successfully recovered, under varying latent score gap $\Delta_K$.

\
\end{document}